\documentclass{article}

\usepackage[preprint]{corl_2026}

\usepackage{amsmath, amssymb, amsfonts, mathtools}
\usepackage{graphicx}
\usepackage{booktabs}
\usepackage{multirow}
\usepackage{subfigure}
\usepackage{wrapfig}
\usepackage{float}
\usepackage[table]{xcolor}

\usepackage[ruled, vlined]{algorithm2e}

\usepackage{tikz}
\usetikzlibrary{positioning, arrows.meta, shapes.geometric, fit, backgrounds}
\definecolor{hgreen}{RGB}{0,100,60}
\definecolor{hblue}{RGB}{0,51,102}
\definecolor{horange}{RGB}{180,90,0}

\title{SSC: A Verifiable Structured Representation \\
       for Bimanual Manipulation Labelling}

\author{
  Yupu Lu \\
  School of Computing and Data Science \\
  The University of Hong Kong, HKSAR \\
  \texttt{luyp16@connect.hku.hk} \\
  \And
  Shuang Wu \\
  Artificial Intelligence Laboratory (Leibniz), HKSAR \\
  \texttt{wushuangust@gmail.com} \\
  \And
  Sihan Chen, Ruihua Han \\
  School of Computing and Data Science \\
  The University of Hong Kong, HKSAR \\
  \And
  Yichen Zhang, Marcus Kalander \\
  Artificial Intelligence Laboratory (Leibniz), HKSAR \\
  \And
  Jia Pan \\
  School of Computing and Data Science \\
  The University of Hong Kong, HKSAR \\
  \texttt{panj@connect.hku.hk}
}

\begin{document}
\maketitle

\begin{abstract}
Subtask labels decompose a long-horizon manipulation demonstration into shorter semantic segments for policy training and evaluation.
Natural-language descriptions are easy to read, but their linguistic variability makes automatic verification difficult.
Rigid template formats, such as BEHAVIOR-1K's \texttt{skill\_annotation}, are linguistically over-segmented, hindering both readability and annotation consistency.
We propose the \textbf{Structured Subtask Chain (SSC)}, a state-transition representation that bridges these extremes.
A demonstration is a sequence of \textbf{Structured Subtask Template (SST)} entries. Each SST stores core action components (subject, predicate, object), flexible conditions (adverbial modifiers such as spatial or instrumental phrases), a base-motion field separate from arm actions, and an after-state scene graph.
Built on this format, SSC supports three vision-language (VL)–assisted functions: rendering SSTs as natural language, checking the assembled chain against four state-transition rules, and completing underspecified fields through a query resolution cascade.
We instantiate the pipeline on BEHAVIOR-1K (50 tasks, 3 episodes per task, 2{,}357 annotated action cells) for logic verification and content completion, evaluating 13 selected state-of-the-art VL models as candidate verifiers and reporting labelling anomalies.
\end{abstract}

\keywords{Structured Subtask Chain, Structured Subtask Template, Subtask Labelling, Vision-Language Verifier, Bimanual Manipulation, BEHAVIOR-1K}

\section{Introduction}
\label{sec:intro}

Large-scale imitation learning for robotic manipulation, especially recent vision-language-action (VLA) generalist policies such as $\pi_{0.7}$~\citep{intelligence2026pi} or \texttt{LingBot-VLA}~\citep{wu2026pragmatic}, increasingly depend on demonstration datasets pre-segmented into dense, semantically-labelled subtasks.
The representation of those labels determines what downstream consumers can do with them: hierarchical policies need structured arm and object fields, VLA training-data curators need labels that admit scalable automatic checks, and human reviewers need labels that compose back into readable task descriptions.

Two families of subtask-label representations dominate today, and each fails on at least one of these axes.
\textbf{Natural-language descriptions} are readable, but have no structural invariant and are hard to verify autimatically because of their linguistic variability~\citep{jiang2025galaxea}.
\textbf{Rigid template formats}, such as BEHAVIOR-1K's \texttt{skill\_annotation}~\citep{li2023behavior}, are structured but suffer from two parallel issues: over-segmentation into many narrow fields and complex linguistic logic, which cause verification issues and leave key components unfilled in some tasks. Together these issues affect both readability and cross-task consistency.

We propose the \textbf{Structured Subtask Chain (SSC)} (Sec.~\ref{sec:formalism}), a state-transition representation that occupies the structured-yet-flexible middle ground: a demonstration is a chain of \textbf{Structured Subtask Template (SST)} entries, each pairing grammatical action fields and flexible modifiers, along with separate base motion and an after-state scene graph describing current situation.

Around this representation we build a three-stage labelling pipeline (Sec.~\ref{sec:method}): \textbf{(i)~SST construction} records unresolved fields as typed pending queries; \textbf{(ii)~a three-level cascade} resolves them by trajectory inference, opportunistic annotation lookup, and typed VL queries; and \textbf{(iii)~consistency checking} validates the assembled chain against four state-transition invariants (hold/release symmetry, object-identity continuity, gripper-event--window alignment, and bimanual hand-over legality), providing automatic invariant-based verification of the chain.

We evaluate SSC on BEHAVIOR-1K as a case study (Sec.~\ref{sec:experiments}) over 50 household tasks, 3 episodes per task, and 13 VL models from five providers, reporting per-model verifier accuracy, task performance, and anomalies raised by the consistency checker.

\section{Related Work}
\label{sec:related}

\paragraph{Subtask-Conditioned Policies for Long-Horizon Manipulation.}
Recent vision-language-action (VLA) foundation models, of which $\pi_{0.7}$~\citep{intelligence2026pi} and \texttt{LingBot-VLA}~\citep{wu2026pragmatic} are a prominent example, have rapidly adopted subtask decomposition for general manipulation, targeting on long-horizon tasks~\citep{liu2026long, larchenko2025behavior}.
A parallel keypose-policy line predicts explicit sub-goals before low-level trajectory generation~\citep{ma2024hierarchical, xian2023chaineddiffuser, yu2024bikc, xu2025bikc+}.
These works all motivate structured subtask reasoning, which our pipeline makes explicit and checkable, including arm assignment for bimanual demonstrations.

\paragraph{Manipulation Datasets and the Annotation Bottleneck.}
Manipulation datasets cover simulated platforms~\citep{james2020rlbench, mu2021maniskill, li2023behavior}, real-robot teleoperation~\citep{fang2024rh20t, wu2025robomind, jiang2025galaxea}, human bimanual demonstrations~\citep{kuehne2014language, zhan2024oakink2, carmona2025bicap}, and instruction-following benchmarks with long-horizon language-conditioned episodes~\citep{shridhar2020alfred, padmakumar2022teach}.
Their annotations expose heterogeneous structure, such as language descriptions, subtask-level language, task hierarchies, action grammars, or dialogues, largely operating at the task or cross-subtask level. The SSC representation introduced in this paper sits at a complementary, within-subtask level, helping oragnize and verify the subtask labels in a template but maintain contents flexibility. 

\paragraph{Subtask Labelling: VL Annotators and Temporal Action Segmentation.}
Two types of approaches have produced complementary labelling tools.
\emph{VL-based annotators} leverage pretrained VL models to propagate language onto unlabelled demonstrations or jointly identify keyframes with skill language~\citep{xiao2023dial, kuramshin2025task, kou2025roboannotatorx, kou2024kisa}.
\emph{Temporal action segmentation} treats labelling as a discriminative video task with multi-stage temporal convolutions, query transformers, and supporting annotation interfaces~\citep{farha2019ms, wang2026timestamp, ding2023temporal, stanovcic2026atlas}.
Our SSC format complements both lines with a structured state-transition representation and a four-invariant consistency check (Sec.~\ref{sec:ssc-validation}).

\section{Framework Formalism}
\label{sec:formalism}

We model a manipulation demonstration as a \emph{state-transition system}, similar to the state-transition view in world-model design~\citep{hao2023reasoning}. A demonstration is an alternating sequence of states and actions,
\begin{equation*}
    g_0 \xrightarrow{A_1} g_1 \xrightarrow{A_2} g_2 \xrightarrow{A_3} \cdots \xrightarrow{A_{K-1}} g_{K-1},
\end{equation*}
where a state $g_k$ is a scene-graph snapshot recording which objects are held by which arm, and an action set $A_k$ describes the structural change between consecutive states. With this state-transition view in place, the pipeline annotates each transition with structured labels and resolves any residual ambiguity by progressively richer information sources. The full Python schema is given in Appendix~\ref{app:schema}.

\subsection{Structured Subtask Template and Chain}
\label{sec:sst}

An \emph{Structured Subtask Template} (SST) entry $S_k$ is a record carrying everything required to characterise one contiguous interval of the demonstration:
\begin{equation}
    S_k \;=\; (\,\mathrm{duration}_k,\; \mathrm{motion}_k,\; A_k,\; g_k,\; Q_k\,),
\end{equation}
where $\mathrm{duration}_k = (s_k, e_k]$ is a half-open frame range, $\mathrm{motion}_k \in \{\texttt{idle}, \texttt{move}\}$ records whether the robot base translates during the interval, $A_k$ is an ordered list of action entries describing the transition from $g_{k-1}$ to $g_k$, $g_k$ is the scene-graph snapshot of the \emph{after-state} of $S_k$, $Q_k$ is a list of pending queries that record any residual ambiguity in the entry (see Sec.~\ref{sec:resolution}), and
$S_0 = (\,[0,0],\,\texttt{idle},\,\emptyset,\,g_0,\,\emptyset\,)$ is a trivial state.

The \emph{Structured Subtask Chain} (SSC) is the dense sequence $\{S_k\}_{k=0}^{K-1}$.
The induced sequence of after-states $g_0,\, g_1,\, \ldots,\, g_{K-1}$ is the \emph{state trajectory} of the demonstration, and is the quantity our pipeline maintains as a running variable across the entire labelling process.

\paragraph{Grammatical decomposition of an action entry.}
Each action entry $a \in A_k$ decomposes one observed action into grammatical components:
a \emph{subject} (\texttt{arm} $\in$ \{\texttt{left}, \texttt{right}, \texttt{coop}, \texttt{unsure}\}),
a \emph{predicate} (an \texttt{action\_type} $\in$ \{$\mathtt{grasp}$, $\mathtt{tool\_use}$, $\mathtt{contact}$, $\mathtt{idle}$\} + an \texttt{action\_verb}),
an \emph{object} (\texttt{action\_obj}),
and \emph{conditions} (adverbial modifiers such as spatial and instrumental modifiers on the action and on the object, e.g.\ \texttt{from breakfast\_table} or \texttt{with knife}; conditions are intentionally flexible and admit several correct phrasings around the same physical action).
The structured fields compose directly into a natural-language sentence (\emph{``right arm picks book from breakfast table''}), so an SSC can be reconstructed as a human-readable transcript or audited field-by-field as machine-readable structure.

\paragraph{Motion is separate from arm-side actions.}
$\mathrm{motion}_k$ is kept outside the action list because base translation and arm manipulation can occur independently and are inferred from disjoint signals. A subtask in which the robot drives forward while one arm holds an object can merge the two streams into one single $\texttt{move~and~hold}$ phrase in the natural-language description.

\paragraph{Scene-Graph State Space. } 
A scene-graph snapshot $g$ is a triple
\begin{equation}
    g = (\,\mathrm{left}(g),\; \mathrm{right}(g),\; \mathrm{same}(g)\,),
\end{equation}
where $\mathrm{left}(g), \mathrm{right}(g) \in \mathcal{O} \cup \{\bot\}$ are the object identifiers held by each hand (with $\bot$ denoting an empty hand) and $\mathrm{same}(g) \in \{\mathsf{T}, \mathsf{F}, \mathsf{?}\}$ records whether the two hands hold the \emph{same} physical object (which can be true, false, or unresolved when both hands hold same-named objects whose identity cannot be inferred from kinematics alone).
Up to renaming of objects, the state space contains exactly seven configurations (Table~\ref{tab:state-configs}).

\begin{table}[h]
\centering
\caption{The seven scene-graph configurations. $\mathsf{D_?}$ is the only configuration with unresolved state; the resolution cascade of Sec.~\ref{sec:resolution} promotes $\mathsf{D_?}$ to either $\mathsf{D_d}$ or $\mathsf{S}$ by a VL query.}
\label{tab:state-configs}
\small
\begin{tabular}{lllll}
\toprule
Config & $\mathrm{left}(g)$ & $\mathrm{right}(g)$ & $\mathrm{same}(g)$ & Description \\
\midrule
$\mathsf{E}$  & $\bot$        & $\bot$         & $\mathsf{F}$ & Both hands empty \\
$\mathsf{L}$  & $o_A$         & $\bot$         & $\mathsf{F}$ & Only left holds \\
$\mathsf{R}$  & $\bot$        & $o_B$          & $\mathsf{F}$ & Only right holds \\
$\mathsf{D}$  & $o_A$         & $o_B \neq o_A$ & $\mathsf{F}$ & Different objects \\
$\mathsf{D_d}$& $o_A$         & $o_A$ (instance)&$\mathsf{F}$ & Same-named, confirmed different instances \\
$\mathsf{S}$  & $o_X$         & $o_X$          & $\mathsf{T}$ & Same physical object (co-grasp) \\
$\mathsf{D_?}$& $o_A$         & $o_A$          & $\mathsf{?}$ & Same-named, identity unresolved \\
\bottomrule
\end{tabular}
\end{table}

\section{Methodology}
\label{sec:method}

\subsection{Pipeline Overview}
\label{sec:overview}

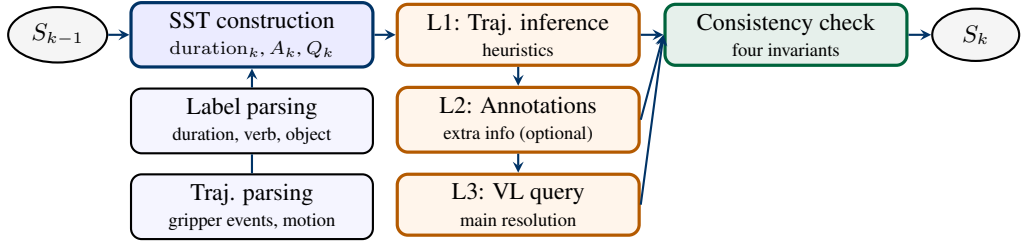
\begin{figure}[h]
\centering
\begin{tikzpicture}[
  node distance=0.4cm and 0.28cm,
  font=\footnotesize,
  box/.style={draw, thick, rounded corners, align=center,
              minimum width=3.2cm, minimum height=0.7cm, fill=blue!4},
  rbox/.style={draw=hgreen, very thick, fill=hgreen!10, rounded corners,
               minimum width=3.2cm, minimum height=0.7cm, font=\footnotesize, align=center},
  bbox/.style={draw=hblue, very thick, fill=hblue!8, rounded corners,
               minimum width=3.2cm, minimum height=0.7cm, font=\footnotesize, align=center},
  cascade/.style={draw=horange, very thick, rounded corners, align=center,
                  minimum width=3.2cm, minimum height=0.7cm, fill=orange!8},
  state/.style={draw, ellipse, align=center, minimum width=1.15cm, fill=gray!8},
  arr/.style={->, >=stealth, hblue, thick},
  drr/.style={-, >=stealth, hblue, thick},
  fwd/.style={->, >=stealth, gray!70!black, thick, dashed}
]
  \node[state, thick] (sk-1) {$S_{k-1}$};
  \node[bbox, right=of sk-1, fill=blue!8] (sst) 
        {SST construction\\{\scriptsize $\mathrm{duration}_k$, $A_k$, $Q_k$}};
  \node[box, below=0.25cm of sst] (anno) 
        {Label parsing\\{\scriptsize duration, verb, object}};  
  \node[box, below=0.25cm of anno] (traj) 
        {Traj. parsing\\{\scriptsize gripper events, motion}};
  \node[cascade, right=of sst] (l1) 
        {L1: Traj. inference \\{\scriptsize heuristics}};
  \node[cascade, below=0.25cm of l1, fill=orange!8] (l2) 
        {L2: Annotations\\{\scriptsize extra info (optional)}};
  \node[cascade, below=0.25cm of l2] (l3) 
        {L3: VL query\\{\scriptsize main resolution}};
  \node[rbox, right=0.3cm of l1] (check)
        {Consistency check\\{\scriptsize four invariants}};
  \node[state, right=0.3cm of check, thick] (sk) {$S_k$};

  \draw[arr] (sk-1.east) -- (sst.west);
  \draw[arr] (anno.north) -- (sst.south);
  \draw[drr] (traj.north) -- (anno.south);
  \draw[arr] (sst.east) -- (l1.west);
  \draw[arr] (l1.south) -- (l2.north);
  \draw[arr] (l2.south) -- (l3.north);
  \draw[arr] (l1.east) -- (check.west);
  \draw[arr] (l2.east) -- (check.west);
  \draw[arr] (l3.east) -- (check.west);
  \draw[arr] (check.east) -- (sk.west);
\end{tikzpicture}
\caption{Per-subtask processing flow. For each subtask interval, SST construction parses the annotation and trajectory into an entry $S_k$ with pending queries $Q_k$; the cascade resolves $Q_k$ by Level~1 (trajectory inference), optionally Level~2 (annotation lookup) when dataset metadata supports it, and Level~3 (VL queries) for residual ambiguity; the consistency checker validates $S_k$ against four state-transition invariants; the resolved after-state $g_k$ in $S_k$ then becomes the before-state for $S_{k+1}$.}
\label{fig:workflow}
\end{figure}

The SSC pipeline runs in three stages (Fig.~\ref{fig:workflow}).
\emph{(i)~SST construction} (Sec.~\ref{sec:sst-construction}) parses each subtask annotation and the underlying trajectory into an SST entry, recording any fields that cannot be determined at this stage as typed pending queries.
\emph{(ii)~The three-level resolution cascade} (Sec.~\ref{sec:resolution}) settles those queries by preferring trajectory inference over annotation lookup and annotation lookup over vision-language (VL) queries.
\emph{(iii)~Consistency checking} (Sec.~\ref{sec:ssc-validation}) validates the assembled chain against four state-transition invariants.
Stages~(i) and~(ii) are interleaved in an online architecture: each SST entry is fully resolved before the next is built, so the resolved scene-graph state at entry $k$ propagates into the trajectory inferences applied to entry $k+1$.

\subsection{SST Construction}
\label{sec:sst-construction}

We assume each subtask annotation interval contains at most one stateful action per arm; intervals containing more than one sequential stateful action by the same arm must be split upstream. Each annotation then passes through three steps that together populate every field of $S_k$.

\emph{Label parsing} extracts the action verb, the action type, the action object, any spatial or instrumental conditions, and the acting arm when the label states it explicitly; the parser patterns are dataset-specific to the labelling style.

\emph{Trajectory parsing} extracts per-arm gripper open/close events from the gripper-command stream and classifies each subtask's base motion as \texttt{idle} or \texttt{move} from the smoothed lower-body signal.

\emph{SST construction} assembles the entry in two steps. First, it advances the running scene-graph state $g_k$ under the parsed action case: persistent acquire/release, stateless contact, or transient manipulation (whose temporary hold is cleared before the next entry). Second, it records every field that the parsed inputs cannot uniquely determine into $Q_k$ as a typed query ($\mathtt{arm\_assignment}$, $\mathtt{same\_object}$, or $\mathtt{empty\_hand}$), each flagged with whether VL resolution is required.

Related verb vocabulary, the parsed action logic, and the typed query schema are detailed in Appendix~\ref{app:impl-details}.

\subsection{State Resolution Cascade}
\label{sec:resolution}

SST construction emits each entry $S_k$ with most fields populated but with $Q_k$ recording any residual ambiguity. The cascade resolves $Q_k$ in three levels of increasing cost.

\textbf{Level~1 (L1): Trajectory inference.}
L1 combines trajectory signals with the running scene-graph state to settle queries that can be resolved or verified from trajectory data. In the current implementation, L1 targets $\mathtt{arm\_assignment}$ queries on grasp and release transitions with gripper events.
A single arm's gripper close or open event identifies that arm as the acting one and confirms \emph{which arm holds, or just released, which object}; a simultaneous bimanual gripper event identifies both arms as acting. L1 also recognises a \emph{cooperative release} from the scene-graph state alone, a release of an object that was previously co-grasped, and aligns action entry to reflect this cooperative behavior.

\textbf{Level~2 (L2, optional): Annotation lookup.}
When the dataset supplies extra metadata that can disambiguate same-object identity, for example, type tags on the action, prefixes on object names, or instance-level object identifiers, L2 consults this metadata to resolve queries before reaching the VL layer. 
This level is optional: it is skipped when the dataset lacks reliable metadata for the queried field, and on a per-field basis when the metadata signal is unreliable.

\textbf{Level~3 (L3): Vision-language queries.}
L3 handles residual queries that L1 and any available L2 metadata cannot resolve. The query kind determines the prompt template: an $\mathtt{arm\_assignment}$ query asks ``which arm performs this action?'', and a $\mathtt{same\_object}$ query asks ``are the two hands on the same physical object or two same-named instances?''. The full template text and JSON response schema are given in Appendix~\ref{app:prompts} and can be extended to new query types as needed. 

\paragraph{Ordering and online execution.}
Within each SST entry, the cascade resolves pending queries in dependency order. 
It first answers queries that determine the scene graph, then answers queries whose prompt depends on that fixed before- or after-state. Construction and cascade execution are interleaved per entry. 

After $Q_k$ is resolved, $g_k$ becomes the before-state of $S_{k+1}$, so resolved decisions made at $S_k$ can support deterministic inference at the next entry and avoid re-raising downstream queries. The snapshot/re-advance mechanic that implements this online interleaving is described in Appendix~\ref{app:impl-details}. Fig.~\ref{fig:traced-example} illustrates the workflow on a three-subtask trace where a single L3 resolution at an acquire subtask deterministically settles downstream release of the same object via L1.

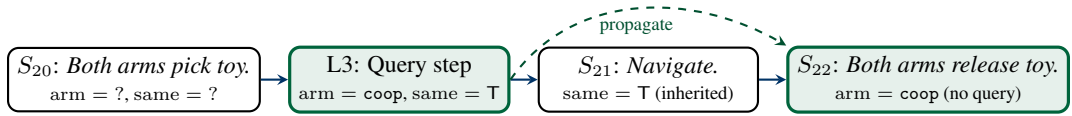
\begin{figure}[h]
\centering
\begin{tikzpicture}[
  node distance=0.25cm,
  box/.style={draw, thick, rounded corners, minimum width=2.9cm, minimum height=0.55cm,
              font=\footnotesize, align=center},
  rbox/.style={draw=hgreen, very thick, fill=hgreen!10, rounded corners,
               minimum width=2.9cm, minimum height=0.55cm, font=\footnotesize, align=center},
  arr/.style={->, >=stealth, hblue, thick},
  garr/.style={->, >=stealth, hgreen!80!black, thick, dashed}
]
  \node[box] (s20) {$S_{20}$: \emph{Both arms pick toy.}\\{\scriptsize $\mathrm{arm}=\mathsf{?}$, $\mathrm{same}=\mathsf{?}$}};
  \node[rbox, right=0.36cm of s20] (ann) 
  {L3: Query step\\{\scriptsize $\mathrm{arm}=\texttt{coop}$, $\mathrm{same}=\mathsf{T}$}};
  \node[box, right=0.36cm of ann] (s21) {$S_{21}$: \emph{Navigate.}\\{\scriptsize $\mathrm{same}=\mathsf{T}$ (inherited)}};
  \node[rbox, right=0.36cm of s21] (s22) {$S_{22}$: \emph{Both arms release toy.}\\{\scriptsize $\mathrm{arm}=\texttt{coop}$ (no query)}};
  \draw[arr] (s20) -- (ann);
  \draw[arr] (ann) -- (s21);
  \draw[arr] (s21) -- (s22);
  \draw[garr] (ann.east) to[out=55,in=150]
    node[below, font=\scriptsize, hgreen!80!black] {propagate} (s22.north west);
\end{tikzpicture}
\caption{A three-subtask trace of the online cascade. \textbf{Setup:} at $S_{20}$ both arms acquire a same-named object; trajectory and annotation alone cannot tell whether they hold one physical instance ($\mathsf{S}$) or two ($\mathsf{D_d}$). \textbf{Resolution:} the cascade falls through to Level~3, issues a $\mathtt{same\_object}$ query, and commits $\mathrm{same}=\mathsf{T}$ as the after-state of $S_{20}$. \textbf{Consequence:} the resolved state propagates through $S_{21}$ and into the before-state of $S_{22}$, so Level~1 deterministically infers a cooperative release ($\mathrm{arm}=\texttt{coop}$) at $S_{22}$ with no further VL call. A pipeline that resolved each subtask in isolation would re-raise the same-object query at $S_{22}$.}
\label{fig:traced-example}
\end{figure}

\subsection{Subtask Chain Consistency Checking}
\label{sec:ssc-validation}

An SSC carries enough state information to validate its own consistency. The assembled chain is checked against four state-transition invariants, and any violation is flagged as an anomaly rather than silently overwritten:
\textbf{(I1)~Hold/release symmetry} --- every $\mathtt{grasp}$ acquisition of an object by an arm must be matched by a subsequent release of the same object by the same arm, or by a cooperative release of an object acquired via $\mathrm{arm}=\texttt{coop}$;
\textbf{(I2)~Object-identity continuity} --- a held object persists into the next entry unless a release-class verb targets the holding arm;
\textbf{(I3)~Gripper-event--window alignment} --- for every grasp/release action, the gripper close/open event that physically realises it must fall inside the action's annotated frame range $(s_k, e_k]$; an event lying outside the window flags an annotation-timing anomaly;
\textbf{(I4)~Bimanual hand-over legality} --- a $\mathtt{hand\_over}$ requires a release event on the giver arm and a close event on the receiver arm within the same window, with the object moving from the giver's slot to the receiver's slot.
The transition tables that ground these invariants are given in Appendix~\ref{app:transitions}, and Appendix~\ref{app:completeness} establishes that the I1--I4 set is complete under the framework's stated assumptions, extendable to future development as needed.

These four invariants operate on the state trajectory rather than on specific verbs. In Sec.~\ref{sec:experiments}, we therefore report two complementary measures: a \emph{local per-action accuracy} on the (\texttt{arm}, \texttt{action\_type}, \texttt{action\_verb}, \texttt{action\_obj}) tuple, which scores label-by-label correctness against a reference SSC; and a \emph{global SSC anomaly count} that totals I1--I4 violations across the chain (evaluated per arm and per entry, yielding a vector of anomaly counts indexed by invariant type). The two are independent: an SSC can be locally accurate but globally inconsistent (e.g.\ a release that does not match any acquire) or globally consistent but locally wrong (e.g.\ a systematic misnaming of one object), so reporting both surfaces failure modes that either metric alone would miss.

\section{Experiments}
\label{sec:experiments}

We conduct a case study to (i)~present how the pipeline works end-to-end, (ii)~comprehensively check the anomalies it surfaces in the human annotations, and (iii)~evaluate the performance of VL models as L3 verifiers.

\subsection{Dataset and Evaluation Protocol}
\label{sec:protocol}

We evaluate the pipeline on BEHAVIOR-1K~\citep{li2023behavior}, a simulated bimanual platform (R1Pro with two 7-DOF arms, mobile base, and torso) recorded at 30\,Hz over 50 household manipulation tasks; the dataset's annotation schema and the per-cascade-level input mapping are summarised in Appendix~\ref{app:b1k-structure}.
For each task we score first three episodes (\texttt{ep0}, \texttt{ep1}, \texttt{ep2}), yielding 2{,}357 annotated action cells in total (786, 791, and 780 action cells per episode respectively).

\paragraph{Operating mode.}
We adopt the metadata-minimal VL-only mode (cascade Level~1 $\to$ Level~3) as the main reporting mode: the pipeline then relies only on the trajectory and the VL verifier, which requires no extra metadata and isolates the VL model's contribution. In parallel, the annotation-augmented VL (AVL) mode (Level~1 $\to$ Level~2 $\to$ Level~3), which additionally consumes BEHAVIOR-1K's structured fields (\texttt{skill\_type}, \texttt{memory\_prefix}, \texttt{spatial\_prefix}) when available, is conducted as an ablation that quantifies the lift reliable annotations can buy when present; full results are in Appendix~\ref{app:avl-ablation}.

\paragraph{Reference annotation.}
For each (task, episode, action) cell we build a reference label in two stages. (i)~A panel of VL models evaluates the cell and casts one vote per model on the cell's arm and same-object assignment; the majority answer across the panel is recorded as a candidate label. (ii)~A human reviewer inspects every candidate against the recorded video and overrides it when the panel majority disagrees with the visible action. The resulting annotated reference SSC is viewed as the ground truth against which all systems are scored. Within this reference we define the \emph{VL trimmed-hard} subset of 138 cells, comprising 134 panel-disagreement cells from (i) and 4 retained-ambiguity cells from (ii). We view this subset as the hard tier used throughout Sec.~\ref{sec:eval-per-model}--\ref{sec:eval-ensemble}. The corresponding 56-cell AVL counterpart is defined in Appendix~\ref{app:avl-ablation}.

\paragraph{Models.}
We test 13 VL models in both modes; each is referred to by a short tag throughout the paper:
Alibaba
\texttt{qwen3-vl-plus (qwen3vlp)},
\texttt{qwen3-vl-flash (qwen3vlf)},
\texttt{qwen-vl-max (qwenvlx)},
ByteDance
\texttt{doubao-seed-2.0-pro (dbao2p)},
Google
\texttt{gemini-2.5-pro (gem25p)},
\texttt{gemini-2.5-flash (gem25f)},
\texttt{gemini-3.1-pro-preview (gem31p)},
\texttt{gemini-3.5-flash (gem35f)},
Moonshot
\texttt{kimi-k2.5 (kimi25)},
OpenAI
\texttt{gpt-5.4 (gpt54)},
\texttt{gpt-5.4-mini (gpt54mi)},
\texttt{gpt-5.5 (gpt55)},
\texttt{o4-mini (o4mi)}.

Each (task, episode, model, mode) tuple is run once, totalling 3{,}900 successful runs (50 tasks $\times$ 3 episodes $\times$ 13 models $\times$ 2 modes) with zero failures.

\subsection{Per-Model Action-Level Accuracy}
\label{sec:eval-per-model}

Table~\ref{tab:per-model} reports the overall and hard-tier action-level accuracy of each model under the VL-only operating mode. The AVL ablation test is reported in Appendix~\ref{app:avl-ablation}.

\begin{table}[h]
\centering
\caption{Per-model VL-only action-level accuracy on BEHAVIOR-1K (50 tasks $\times$ 3 episodes, 2{,}357 annotated action cells). Models are ordered left to right by trimmed-hard accuracy on the 138-cell VL hard subset; best per row in bold.}
\label{tab:per-model}
\scriptsize
\setlength{\tabcolsep}{1.5pt}
\begin{tabular}{@{}l|ccccccccccccc@{}}
\toprule
\textbf{Model} & \texttt{gpt55} & \texttt{gem35f} & \texttt{gpt54} & \texttt{gem31p} & \texttt{dbao2p} & \texttt{gem25p} & \texttt{qwen3vlf} & \texttt{kimi25} & \texttt{gem25f} & \texttt{qwenvlx} & \texttt{gpt54mi} & \texttt{o4mi} & \texttt{qwen3vlp} \\
\midrule
Overall (\%)      & \textbf{99.19} & 99.15 & 99.11 & 99.07 & 99.02 & 99.02 & 98.90 & 98.73 & 98.51 & 98.13 & 98.01 & 98.01 & 97.84 \\
Hard/138 (\%)     & \textbf{86.23} & 85.51 & 84.78 & 84.78 & 83.33 & 83.33 & 81.16 & 78.26 & 74.64 & 68.12 & 65.94 & 65.94 & 65.94 \\
\bottomrule
\end{tabular}
\end{table}

The leaderboard is led by \texttt{gpt55} (86.23\%) and \texttt{gem35f} (85.51\%), with \texttt{gpt54} and \texttt{gem31p} (both 84.78\%) close behind, and \texttt{dbao2p} and \texttt{gem25p} tied at 83.33\%. The top six models lie within $\sim$3\,pp of one another on the 138-cell hard subset, so single-model selection is unlikely to be the dominant lever for further improvement. At the bottom of the column, \texttt{qwen3vlp}, \texttt{gpt54mi}, and \texttt{o4mi} all sit at 65.94\%, about 20\,pp below the leader.

\subsection{Per-Task Difficulty Tiers}
\label{sec:eval-tiers}

Per-model accuracy aggregates across 50 tasks; we now drill into where the residual hard cells live on the 138-cell VL trimmed-hard subset, with detailed tables in Appendix~\ref{app:difficulty-tiers}.

\emph{(i)~Global vote-margin view.} The trimmed-hard subset is anchored to the original 13-model panel; on its 134 genuine disagreements, the pool's vote margin (majority minus runner-up, out of 13) splits into \emph{decisive} (margin $\geq 6$, 93 cases, 69\%), \emph{intermediate} (margin 3--5, 29 cases, 22\%), and \emph{tight} (margin $\leq 2$, 12 cases, 9\%) bands; the intermediate band is the main lever for ensemble selection, and the tight band accounts for most residual ensemble error.

\emph{(ii)~Per-task tier view.} Classifying the same 138 cells by the number of production-pool models that solve them yields a 21-cell hardest cohort across three tiers: a \emph{universal-failure} tier of 6 cells (0/13 solve), a \emph{near-universal-failure} tier of 5 cells (1--2/13), and a \emph{moderate-difficulty} tier of 10 cells (3--6/13); the remaining 117 cells are resolved by a pool majority. One Tier-1 cell has an unresolved reference arm, so per-model totals in Appendix~\ref{app:difficulty-tiers} are reported over 20 scored cells.

The hardest cohort concentrates on bimanual coordinated open / close / push actions on large containers (toolbox, fridge, oven, laptop) --- a single, recognisable failure mode that no model in the 13-model pool solves on any Tier-1 cell and that accounts for the residual $\sim$4--5\% gap the strongest ensembles also cannot close. It is the most concrete improvement target for future verifier work.

\subsection{Multi-VLM Ensemble Exploration}
\label{sec:eval-ensemble}

Canonical multi-model ensembles can further stabilise VL verification. Several top trio candidates balance accuracy and provider diversity on the 138-cell VL trimmed-hard subset, with \texttt{dbao2p + qwen3vlf + gem35f} ($89.13\%$) and \texttt{gpt54 + qwenvlx + gem35f} ($88.41\%$) as representative examples. Full exploration is in Appendix~\ref{app:ensemble}.

\subsection{SSC Logic Completeness and Anomaly Reports}
\label{sec:eval-logic}

We apply the I1--I4 consistency checks of Sec.~\ref{sec:ssc-validation} to the annotated reference SSC across the 150 task--episode pairs (50 tasks $\times$ 3 episodes). The checker flags \textbf{31} action cells out of 2{,}357, distributed as 18 \emph{annotation-timing}, 9 \emph{not-released-at-end}, and 4 \emph{release-without-hold} anomalies; Table~\ref{tab:anomalies} lists the affected tasks and per-episode cell counts.

\begin{table}[h]
\centering
\caption{Per-task anomaly breakdown for the annotated reference SSC over the 50-task evaluation set (3 episodes each). \textbf{n/Ep.} reports cell-level anomaly counts per episode (\texttt{ep0/ep1/ep2}); all rows sum to the headline 18 + 9 + 4 = 31. \textbf{Verdict} marks \emph{out-of-scope} cases (the trajectory violates a framework assumption) and \emph{annotation faults} (the GT label or interval is incorrect).}
\label{tab:anomalies}
\small
\begin{tabular}{cclll}
\toprule
\textbf{Task} & \textbf{n/Ep.} & \textbf{Anomaly kind} & \textbf{Root cause} & \textbf{Verdict} \\
\midrule
0024 & 4/4/4 & annotation\_timing      & Hugging; no gripper-close event (I3)         & Out-of-scope \\
0030 & 1/1/1 & not\_released\_at\_end  & Cooperative hold at episode end (I1)         & Out-of-scope \\
0038 & 1/1/1 & not\_released\_at\_end  & Cooperative hold at episode end (I1)         & Out-of-scope \\
0039 & 1/1/1 & not\_released\_at\_end  & Cooperative hold at episode end (I1)         & Out-of-scope \\
0002 & 0/0/1 & release\_without\_hold  & Misaligned window cascades to release (I1)   & Annotation fault \\
0026 & 1/0/1 & release\_without\_hold  & Misaligned window cascades to release (I1)   & Annotation fault \\
0036 & 0/1/0 & release\_without\_hold  & Misaligned window cascades to release (I1)   & Annotation fault \\
0002 & 0/0/1 & annotation\_timing      & Gripper event outside annotated window (I3)  & Annotation fault \\
0026 & 1/0/1 & annotation\_timing      & Gripper event outside annotated window (I3)  & Annotation fault \\
0033 & 0/1/0 & annotation\_timing      & Gripper event outside annotated window (I3)  & Annotation fault \\
0034 & 1/0/0 & annotation\_timing      & Gripper event outside annotated window (I3)  & Annotation fault \\
0036 & 0/1/0 & annotation\_timing      & Gripper event outside annotated window (I3)  & Annotation fault \\
\bottomrule
\end{tabular}
\end{table}

\begin{figure}[h]
\centering
\begin{minipage}[t]{0.48\textwidth}
\centering
\includegraphics[width=0.32\linewidth]{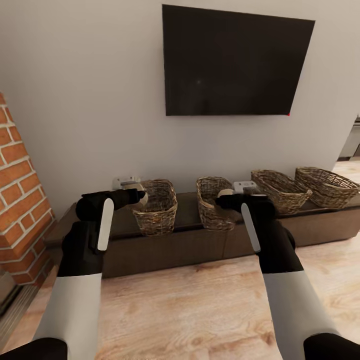}~%
\includegraphics[width=0.32\linewidth]{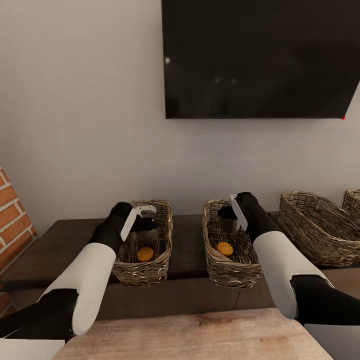}~%
\includegraphics[width=0.32\linewidth]{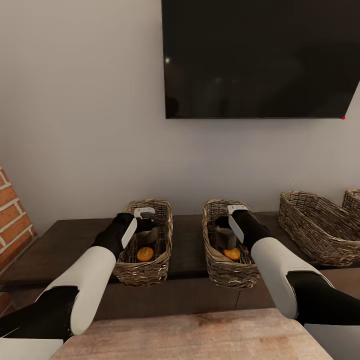}\\[-0.05cm]
{\scriptsize frame 7289 \hspace{1.3cm} 7336 \hspace{1.7cm} 7431}\\[2pt]
\end{minipage}\hfill
\begin{minipage}[t]{0.48\textwidth}
\centering
\includegraphics[width=0.32\linewidth]{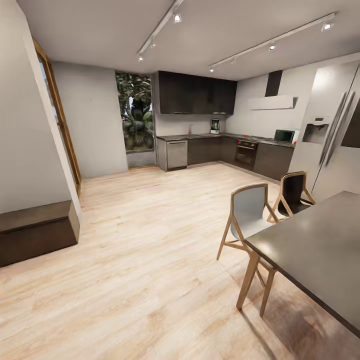}~%
\includegraphics[width=0.32\linewidth]{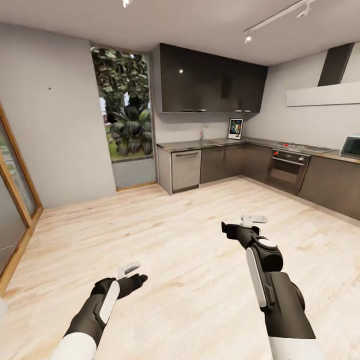}~%
\includegraphics[width=0.32\linewidth]{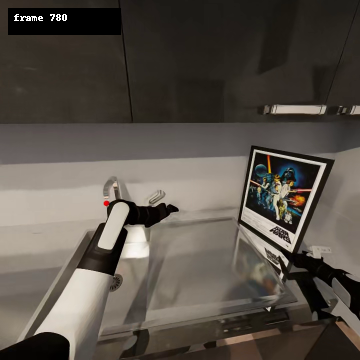}\\[-0.05cm]
{\scriptsize frame 0 \hspace{1.4cm} 235 \hspace{1.8cm} 780}\\[2pt]
\end{minipage}
\caption{Two representative annotation-side anomalies surfaced by the SSC consistency checker. \textbf{Left (Task~26)}: the GT labels two consecutive subtasks over the intervals $[7289,\,7336]$ and $[7336,\,7431]$, but both candle releases actually occur within $[7289,\,7336]$. I3 fires since no gripper event is found in the second subtask. \textbf{Right (Task~34)}: the GT subtask is annotated over $[0,\,235]$ for picking up poster, but the actual grasp event happens around frame~780, far outside the annotated window, so I3 fires. Both anomalies are surfaced for human review rather than silently coerced.}
\label{fig:anomalies}
\end{figure}

\paragraph{Findings.}
Two patterns split the 31 anomalies (Table~\ref{tab:anomalies}): four tasks are \emph{out-of-scope} under assumptions A1 (grasp-only holding) and A2 (empty terminal states), correctly flagged rather than silently coerced (Figure~\ref{fig:out-of-scope} illustrates one of each in Appendix~\ref{app:comp-coverage}); the remaining five tasks contain \emph{annotation faults} where the GT skill window does not contain the gripper event that realises the action, firing I3. Figure~\ref{fig:anomalies} illustrates two annotation-side cases.

\section{Conclusion}
\label{sec:conclusion}

We introduced the Structured Subtask Chain (SSC), a state-transition representation pairing a grammatical decomposition of each bimanual subtask with a scene-graph snapshot, together with a three-stage labelling pipeline (SST construction, a three-level resolution cascade, consistency checking against four state-transition invariants) that supplies automatic invariant-based verification of the assembled chain within the stated assumptions. A case study on BEHAVIOR-1K with 13 VL models shows that VL models can serve as competent L3 verifiers, provider-diverse ensembles add a modest lift, and the consistency checker independently surfaces out-of-scope behaviours and annotation faults that per-action accuracy alone would miss. 

\paragraph{Limitations and future work.}
\label{sec:limitations}
The framework's coverage is bounded by the assumptions in Appendix~\ref{app:comp-assumptions}, so out-of-scope behaviours (e.g.\ gripper-less holding) are flagged rather than silently handled. As a batch feasibility study, this work suggests model-selection guidance for related task families, where consistent strong performance of \texttt{gpt55} and \texttt{gpt54} is observed in future work as well. Paired with a temporal segmentation front end, the same pipeline could extend from \emph{verifier} to \emph{auto-annotator}+\emph{verifier} on datasets without subtask annotations, broadening its applicability.

\clearpage
\acknowledgments{
}

\bibliography{main}

\newpage
\appendix

\section{SSC Data Schema}
\label{app:schema}

The five core record types of the SSC representation are summarised below in Python dataclass notation; the reference implementation lives in \texttt{src/sst.py}. Internal-only fields (resolution flags, propagation tracking) are omitted for brevity.

\begin{small}
\begin{verbatim}
@dataclass
class ActionEntry:
    arm: str                       # "left" | "right" | "coop" | "unsure"
    action_type: str               # "idle" | "grasp" | "contact" | "tool_use"
    action_verb: Optional[str]     # normalised verb stem (e.g., "pick")
    action_obj: Optional[str]      # primary object
    action_obj_cond: List[str]     # object features or part of object
    action_dscp: str               # natural-language description
    action_cond: List[str]         # tools / spatial / instrumental conditions

@dataclass
class SceneGraph:
    obj_in_left_hand:    Optional[str]
    obj_in_right_hand:   Optional[str]
    holding_same_object: Union[bool, str]   # True | False | "unsure"

@dataclass
class PendingQuery:
    kind: str                      # "arm_assignment" | "same_object" | "empty_hand"
    target_action_idx: Optional[int]
    requires_vl: bool
    reason: str                    # human-readable explanation

@dataclass
class SubtaskEntry:                # the SST
    st_duration: Tuple[int, int]   # (start_frame, end_frame)
    st_idx: int                    # 0-based subtask index
    st_motion: str                 # "idle" | "move"
    st_motion_cond: List[str]      # spatial conditions for base motion
    actions: List[ActionEntry]     # 0, 1, or 2 concurrent actions
    actions_gripper: Optional[Dict[str, str]]   # e.g., {"left": "close"}
    scene_graph: SceneGraph        # after-state of this subtask
    pending_queries: List[PendingQuery]

@dataclass
class StructuredSubtaskChain:      # the SSC
    task_name: str
    task_duration: Tuple[int, int]
    object_list: List[str]
    action_dscp_list: List[str]
    subtasks: List[SubtaskEntry]
\end{verbatim}
\end{small}

\section{BEHAVIOR-1K Dataset Structure}
\label{app:b1k-structure}

BEHAVIOR-1K~\citep{li2023behavior} provides simulated demonstrations on the R1Pro bimanual platform (two 7-DOF arms, 1-DOF torso, mobile base) recorded at 30\,Hz across 50 household manipulation tasks; for the case study we use the first three episodes (\texttt{ep0}, \texttt{ep1}, \texttt{ep2}) of each task. Per episode, the dataset provides synchronised RGB streams (head, two wrist cameras, third-person), the full robot proprioception (joint positions, per-arm gripper command, base velocity), and a JSON annotation file containing the per-subtask schedule. Each \texttt{skill\_annotation} entry carries the fields below.

\begin{small}
\begin{verbatim}
"skill_annotation": [ {
    "skill_idx":         int,           # 0-based subtask index
    "frame_duration":    [start, end],  # half-open frame range
    "skill_description": [str, ...],    # NL label, e.g. "left pick bowl"
    "skill_type":        [str, ...],    # canonical action category, e.g. "pick"
    "object_id":         [str, ...],    # instance ids of action target objects
    "memory_prefix":     [str, ...],    # same-object hints, e.g. "the other"
    "spatial_prefix":    [str, ...],    # spatial modifiers, e.g. "near", "on top of"
  }, ... ]
\end{verbatim}
\end{small}

The resolution cascade's three levels consume disjoint slices of this structure: L1 reads the per-arm gripper-command channel from the trajectory to detect open/close events; L2 reads \texttt{skill\_type}, \texttt{memory\_prefix}, and \texttt{spatial\_prefix} when available; L3 reads RGB frames sampled from \texttt{frame\_duration}. The label-parser mapping from \texttt{skill\_description} onto the universal verb table is detailed in Appendix~\ref{app:impl-details}.

\section{State Transitions}
\label{app:transitions}

This appendix enumerates the admissible state transitions between scene-graph configurations (Table~\ref{tab:state-configs}).
The transition set captures the gripper-mediated holding regime: in the present pipeline, every $\mathtt{grasp}$-group acquire or release is triggered by a gripper close or open event, and the consistency-checking module of Sec.~\ref{sec:ssc-validation} validates assembled SSCs against this set.
Configurations that arise from holding without an active gripper grasp are discussed as a limitation in Appendix~\ref{app:comp-assumptions}.

\begin{table}[h]
\centering
\small
\caption{Acquire transitions (gripper close).}
\label{tab:acquire}
\begin{tabular}{lll}
\toprule
Transition & Gripper signal & Actions created \\
\midrule
$\mathsf{E} \to \mathsf{L}$  & left close                                & 1 (arm=left) \\
$\mathsf{E} \to \mathsf{R}$  & right close                               & 1 (arm=right) \\
$\mathsf{E} \to \mathsf{D}$  & both close, different objects             & 2 (arm=left, arm=right) \\
$\mathsf{E} \to \mathsf{S}$  & both close, same physical object          & 1 (arm=coop) \\
$\mathsf{E} \to \mathsf{D_?}$ & both close, same-named obj, unresolved   & 1 (arm=unsure) \\
$\mathsf{L} \to \mathsf{D}$  & right close, different object             & 1 (arm=right) \\
$\mathsf{R} \to \mathsf{D}$  & left close, different object              & 1 (arm=left) \\
$\mathsf{L} \to \mathsf{S}$  & right close, same object (co-grasp)       & 1 (arm=right or coop) \\
$\mathsf{R} \to \mathsf{S}$  & left close, same object (co-grasp)        & 1 (arm=left or coop) \\
\bottomrule
\end{tabular}
\end{table}

\begin{table}[h]
\centering
\small
\caption{Release transitions (gripper open).}
\label{tab:release}
\begin{tabular}{lll}
\toprule
Transition & Gripper signal & Actions created \\
\midrule
$\mathsf{L} \to \mathsf{E}$         & left open  & 1 (arm=left) \\
$\mathsf{R} \to \mathsf{E}$         & right open & 1 (arm=right) \\
$\mathsf{D} \to \mathsf{L}$         & right open & 1 (arm=right) \\
$\mathsf{D} \to \mathsf{R}$         & left open  & 1 (arm=left) \\
$\mathsf{D} \to \mathsf{E}$         & both open  & 2 (arm=left, arm=right) \\
$\mathsf{D_d} \to \mathsf{E}$       & both open  & 2 (confirmed separate instances) \\
$\mathsf{S} \to \mathsf{E}$         & both open  & 1 (arm=coop) \\
$\mathsf{S} \to \mathsf{L}/\mathsf{R}$ & one arm open, co-held object & cooperative release: both arms cleared \\
$\mathsf{D_?} \to \mathsf{E}$ (unresolved) & both open & 1 (arm=unsure) \\
\bottomrule
\end{tabular}
\end{table}

\begin{table}[h]
\centering
\small
\caption{Hand-over, stateless, and tool-release transitions.}
\label{tab:other-transitions}
\begin{tabular}{lll}
\toprule
Transition & Trigger & Actions created \\
\midrule
$\mathsf{L} \to \mathsf{R}$ & left open + right close (simultaneous) & 1 (hand\_over, giver=left) \\
$\mathsf{R} \to \mathsf{L}$ & right open + left close (simultaneous) & 1 (hand\_over, giver=right) \\
self-loop                   & contact verb (push, press, tap, $\ldots$) & 1 (no state change) \\
self-loop                   & tool\_use verb (chop, stir, sweep, $\ldots$) & 1 (no state change) \\
$\mathsf{L} / \mathsf{R} \to \mathsf{E}$ & tool-release verb (insert, hang, attach) & 1 (releases held tool) \\
\bottomrule
\end{tabular}
\end{table}

The state-transition invariants of Sec.~\ref{sec:ssc-validation} are evaluated against this transition set: an SSC is flagged when the observed (or predicted) transition between $g_{k-1}$ and $g_k$ is absent from the tables above, or when the gripper signature recorded in the SST does not match the trigger column of the corresponding row.

\section{Completeness Argument and Scope}
\label{app:completeness}

This appendix establishes that the transition set of Appendix~\ref{app:transitions} is complete \emph{within the framework's stated assumptions}, and enumerates the assumptions themselves so that future relaxations can re-verify completeness against an expanded state space. The argument has three parts: per-configuration enumeration of admissible transitions (Sec.~\ref{app:comp-enum}), a discussion of the coverage boundary and documented out-of-scope cases (Sec.~\ref{app:comp-coverage}), and an explicit assumption table (Sec.~\ref{app:comp-assumptions}).

\subsection{Per-Configuration Transition Enumeration}
\label{app:comp-enum}

For each of the seven scene-graph configurations (Table~\ref{tab:state-configs}) we enumerate every possible next-state under the gripper-event signal alphabet (close left, close right, close both, open left, open right, open both, no change). Cooperative-release transitions from configuration $\mathsf{S}$ are explicitly listed, as is the deferred-resolution behaviour of $\mathsf{D_?}$.

\paragraph{From $\mathsf{E}$ (both empty).}
Five possible next-states: $\mathsf{E} \!\to\! \mathsf{L}$ (left close, 1 action arm=left); $\mathsf{E} \!\to\! \mathsf{R}$ (right close, 1 action arm=right); $\mathsf{E} \!\to\! \mathsf{D}$ (both close, distinct objects, 2 actions); $\mathsf{E} \!\to\! \mathsf{S}$ (both close, same physical object, 1 action arm=coop); $\mathsf{E} \!\to\! \mathsf{D_?}$ (both close, same-named objects with unresolved identity, 1 action arm=unsure). Self-loop on idle yields zero actions.

\paragraph{From $\mathsf{L}$ (left holds).}
Six possible next-states: $\mathsf{L} \!\to\! \mathsf{E}$ (left open, 1 action); $\mathsf{L} \!\to\! \mathsf{D}$ (right close on a distinct object, 1 action arm=right); $\mathsf{L} \!\to\! \mathsf{S}$ (right close on the same object, co-grasp, 1 action arm=right or arm=coop depending on cooperative inference); $\mathsf{L} \!\to\! \mathsf{R}$ (left open + right close, 1 hand-over action giver=left or 1 pick-up action arm=right); self-loop with contact or tool-use verbs (1 stateless action); idle self-loop (0 actions). The simultaneous-both-close case ($\mathsf{L} \!\to\! \mathsf{D'} $ for some $\mathsf{D'}$ that re-occupies the left hand) is excluded by assumption A3 (one object per hand).

\paragraph{From $\mathsf{R}$ (right holds).}
Mirror of $\mathsf{L}$; six possible next-states under the symmetric gripper signals.

\paragraph{From $\mathsf{D}$ (different objects, different names).}
Five possible next-states: $\mathsf{D} \!\to\! \mathsf{R}$ (left open, 1 release action arm=left); $\mathsf{D} \!\to\! \mathsf{L}$ (right open, 1 release action arm=right); $\mathsf{D} \!\to\! \mathsf{E}$ (both open, 2 release actions); self-loop with contact/tool-use verbs on either held object (1 stateless action); idle self-loop.

\paragraph{From $\mathsf{D_d}$ (same-named, confirmed-different instances).}
Same behaviour as $\mathsf{D}$ for releases (\emph{both open}~$\to$~2 actions; single open~$\to$~1 action); the distinction from $\mathsf{D}$ is purely informational (it records that the same-object resolver previously concluded ``different''), so transition semantics are inherited from $\mathsf{D}$ row-by-row.

\paragraph{From $\mathsf{S}$ (same physical object, co-grasp).}
Two stateful next-states plus a cooperative-release branch: $\mathsf{S} \!\to\! \mathsf{E}$ (both open simultaneously, 1 release action arm=coop); $\mathsf{S} \!\to\! \mathsf{L}$ or $\mathsf{S} \!\to\! \mathsf{R}$ (one arm opens) is the \emph{cooperative-release} case --- when the object was acquired via arm=coop, both arms must be cleared regardless of which gripper opens first, so the framework deterministically transitions to $\mathsf{E}$ rather than $\mathsf{L}$ or $\mathsf{R}$. Contact and tool-use verbs on the co-held object remain stateless.

\paragraph{From $\mathsf{D_?}$ (same-named, identity unresolved).}
Resolution-before-transition: the cascade of Sec.~\ref{sec:resolution} promotes $\mathsf{D_?}$ to either $\mathsf{D_d}$ or $\mathsf{S}$ before downstream consumers see it. If the resolver fails (neither L2 annotation lookup nor L3 VL backend produces a decisive same-object answer), the SST keeps $\mathsf{D_?}$ and any release transition emits a single arm=unsure release action; the chain is then flagged at the consistency-checking stage rather than silently coerced to one or the other branch.

\subsection{Coverage Boundary and Open Cases}
\label{app:comp-coverage}

The per-configuration enumeration of Sec.~\ref{app:comp-enum} is exhaustive within the assumptions of Sec.~\ref{app:comp-assumptions}: every $(g_{k-1}, \text{gripper-signal})$ pair admissible under those assumptions has a defined outcome (A1--A6). In verb-level terms, the covered set includes single-arm and bimanual acquire/release, cooperative release from $\mathsf{S}$, hand-over, stateless contact and tool-use actions, tool-release verbs (\texttt{insert}, \texttt{hang}, \texttt{attach}) that clear the held tool, and transient verbs (\texttt{open}, \texttt{close}, \texttt{pull}, \texttt{turn}) whose temporary holds are auto-cleared at the subtask boundary. Violations of asuumptions A1--A5 are not silently coerced but are raised as named anomalies by the consistency checker; A6 (contact is stateless) is enforced by the verb-to-state mapping rather than by an anomaly check. For example, on BEHAVIOR-1K we observe a representative case of each: holding without an active grasp (A1, task~24), non-empty terminal states (A2, tasks~30, 38, 39), partial gripper opening below the open/close threshold (A4, task~26), and gripper events lying outside their annotated subtask interval (A5, task~34); the residue is small and resolvable by a brief manual review pass.

A category should be marked by the present transition set is the \emph{multi-phase transition} within a single annotation interval --- for example, an ``open door'' action can mean "push door to open", but also can decompose into grasp-handle, pull, and release-handle as three sub-phases under one GT skill. We currently model such actions as a single transient verb.

\begin{figure}[h]
\centering
\begin{minipage}[t]{0.48\textwidth}
\centering
\includegraphics[width=0.85\linewidth]{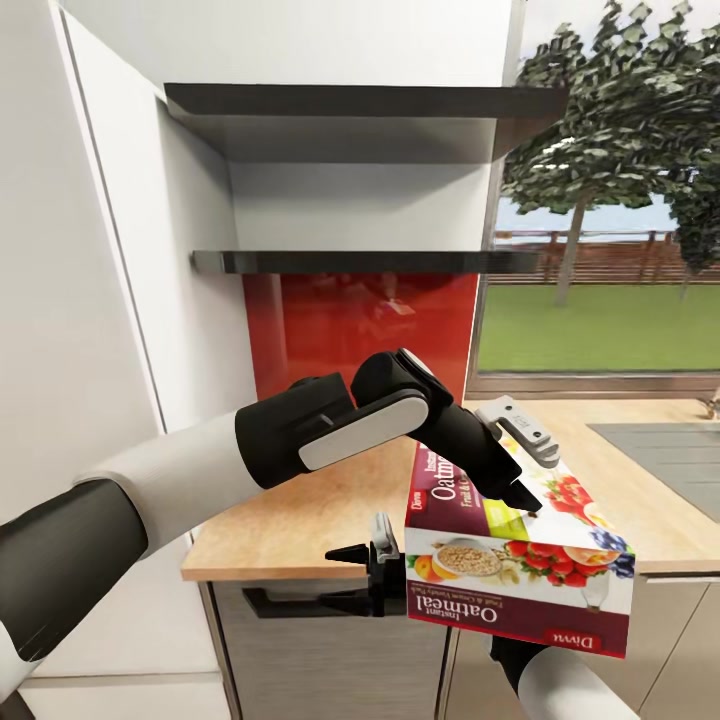}\\[-0.05cm]
{\scriptsize \texttt{task 0024 ep0 st19}, frame 11380}\\[2pt]
\end{minipage}\hfill
\begin{minipage}[t]{0.48\textwidth}
\centering
\includegraphics[width=0.85\linewidth]{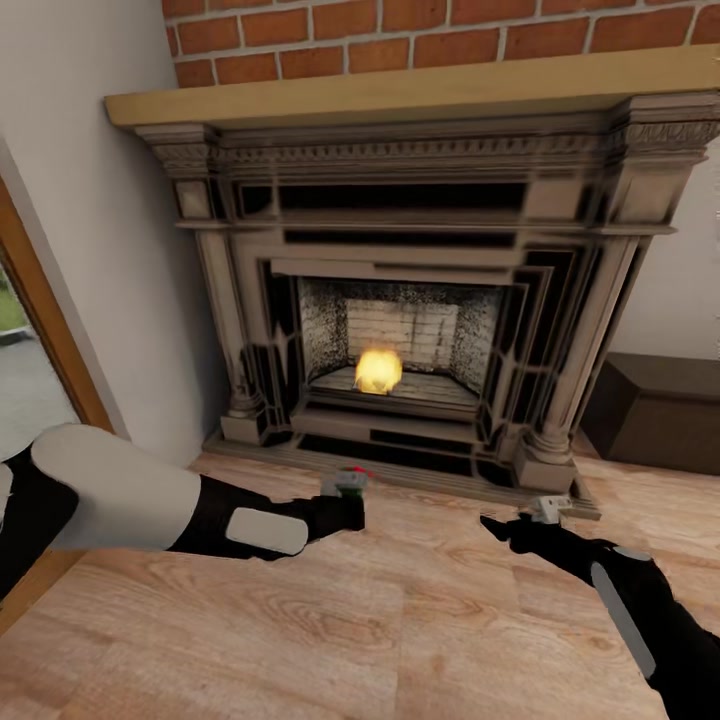}\\[-0.05cm]
{\scriptsize \texttt{task 0030 ep0}, episode-end frame 9650}\\[2pt]
\end{minipage}
\caption{Two illustrative out-of-scope cases surfaced by the consistency checker. \textbf{Left (A1, task 0024)}: the robot \emph{hugs} an oatmeal box between two open grippers without any close event, so no gripper signal places the acquire inside the annotated window and anomaly I3 fires. \textbf{Right (A2, task 0030)}: at the end of the episode the lighter and the holder remain in the grippers, so the trajectory ends with a non-empty terminal state and anomaly I1 fires (a hold never matched by a release).}
\label{fig:out-of-scope}
\end{figure}

\subsection{Assumptions and Their Scope}
\label{app:comp-assumptions}

Each assumption defines a boundary of the framework. Within these boundaries, the transition logic is complete --- every admissible state $\times$ gripper-signal combination has a defined outcome; outside, violations are flagged rather than silently handled. The set of assumptions is fixed for the present version; future relaxations must re-verify completeness on the expanded state space.

\begin{table}[h]
\centering
\caption{Framework assumptions and their relaxation paths. Within these assumptions the transition logic of Appendix~\ref{app:transitions} is complete; outside them, the consistency-checking module of Sec.~\ref{sec:ssc-validation} raises a typed anomaly.}
\label{tab:assumptions}
\small
\begin{tabular}{@{}cp{2.6cm}p{5.4cm}p{4.0cm}@{}}
\toprule
\textbf{ID} & \textbf{Assumption} & \textbf{Completeness impact} & \textbf{Relaxation path} \\
\midrule
A1 & Grasp-only holding & Without this, holding cannot be detected from gripper signals alone. & Add force-based or VL-based holding detection as a fourth signal level. \\
A2 & Empty start and end states & Lets the chain validate the complete transition sequence. & Accept non-empty boundary states and validate against a provided initial state. \\
A3 & One object per arm & Caps the state space at the seven configurations of Table~\ref{tab:state-configs}. & Extend the scene graph to lists per arm; significant complexity increase. \\
A4 & Binary gripper (open vs closed) & Enables clean open/close event detection from the gripper command stream. & Add threshold hysteresis or a continuous gripper-state model. \\
A5 & Action-state coupling & Every grasp action produces a corresponding state change. & Add an ``attempted but failed'' action outcome to the action schema. \\
A6 & Contact is stateless & Contact and tool-use verbs do not modify the holding state. & Already partially relaxed via tool-release verbs (\texttt{insert}, \texttt{hang}, \texttt{attach}). \\
\bottomrule
\end{tabular}
\end{table}

\paragraph{Known contract--implementation gaps.}
The contract above states violations should be flagged rather than silently ignored; the present implementation does not fully satisfy this in three places, which we record as known limitations rather than design choices.
\textbf{(g1)}~An acquire targeting an arm that already holds an object silently overwrites the held object rather than raising an A3/A5 anomaly; a stricter version would emit the anomaly first and require explicit override to continue.
\textbf{(g2)}~Heuristic arm assignment for contact and tool-use actions (the free-arm fallback used when no annotation \texttt{spatial\_prefix} is available) is applied without a low-confidence flag, so downstream consumers cannot distinguish a heuristic decision from a high-confidence one.
\textbf{(g3)}~Base-mode construction uses best-guess running state for arm-side fields that depend on prior holding; if a same-object resolution arrives later, those earlier fields are not re-visited unless they were explicitly marked unsure.
These gaps do not violate the completeness statement above (every transition has a defined outcome), but they mean the boundary between ``confidently resolved'' and ``heuristically resolved'' is not visible to downstream consumers; closing the gaps is a target for future versions.

\section{SST Construction and Cascade}
\label{app:impl-details}

This appendix expands the implementation details that Sec.~\ref{sec:method} cites by reference: the universal verb vocabulary the label parser maps onto, the gripper-event detector, the verb-to-state-transition mapping used during SST construction, the typed pending-query schema, and the snapshot/re-advance mechanic that interleaves the cascade with construction. The BEHAVIOR-1K–specific parser details are isolated in their own paragraph.

\paragraph{Universal verb vocabulary.}
The framework defines four action types, into which any dataset's verbs are mapped by a flat lookup table. The mapping is dataset-agnostic: a new verb is supported by adding one row to the table, without changing any other component of the pipeline. The four types and their canonical members --- as defined in our implementation --- are:
\begin{itemize}\itemsep -0.5ex
\item $\mathtt{idle}$ (base motion or navigation): \texttt{idle}, \texttt{move}, \texttt{move\_to}, \texttt{navigate}, \texttt{navigate\_to}.
\item $\mathtt{grasp}$ (gripper finger movement that acquires, releases, or transiently engages an object), with three sub-roles:
  \emph{acquire} --- \texttt{grasp}, \texttt{pick}, \texttt{pick\_up}, \texttt{pick\_up\_from}, \texttt{pinch}, \texttt{hold};
  \emph{release} --- \texttt{release}, \texttt{place}, \texttt{place\_on}, \texttt{place\_in}, \texttt{place\_on\_next\_to}, \texttt{place\_in\_next\_to}, \texttt{place\_next\_to}, \texttt{place\_under}, \texttt{hand\_over}, \texttt{hand\_over\_to}, \texttt{drop};
  \emph{transient} (may include "engage then release" within the same subtask) --- \texttt{open\_door}, \texttt{close\_door}, \texttt{open\_drawer}, \texttt{close\_drawer}, \texttt{open\_lid}, \texttt{close\_lid}, \texttt{pull}, \texttt{pull\_tray}, \texttt{turn}, \texttt{turn\_to}.
\item $\mathtt{contact}$ (gripper tip directly touches an object without using a held tool; stateless under scene-graph semantics): \texttt{push}, \texttt{push\_to}, \texttt{push\_tray}, \texttt{press}, \texttt{dig}, \texttt{poke}, \texttt{tap}, \texttt{turn\_on}, \texttt{turn\_off}, \texttt{turn\_on\_switch}, \texttt{turn\_off\_switch}, \texttt{tip}, \texttt{tip\_over}.
\item $\mathtt{tool\_use}$ (gripper uses a held object to act on another object), with two sub-roles:
  \emph{persistent} (the tool is retained after the action) --- \texttt{smash}, \texttt{stir}, \texttt{chop}, \texttt{slice}, \texttt{pour}, \texttt{sweep}, \texttt{sweep\_surface}, \texttt{sweep\_off}, \texttt{spray}, \texttt{ignite}, \texttt{wipe}, \texttt{wipe\_hard}, plus the generic \texttt{open}/\texttt{close} when invoked with a held tool;
  \emph{tool-release} (the tool is transferred to the target) --- \texttt{insert}, \texttt{hang}, \texttt{attach}.
\end{itemize}
This four-way division is intentionally narrow and aligned with the scene-graph semantics of Table~\ref{tab:state-configs}; extending the vocabulary to a new dataset typically reduces to adding verbs into one of these four types.

\paragraph{BEHAVIOR-1K label parser.}
On the BEHAVIOR-1K data used in our evaluation, the label parser maps each \texttt{skill\_annotation} string to (verb, object, conditions) by pattern-matching against the verb table above. Where BEHAVIOR-1K attaches instance indices to same-named objects (e.g.\ \texttt{pillar\_candle\_89} versus \texttt{pillar\_candle\_90}), the parser preserves the index so that downstream resolution can distinguish them. A new dataset with a different annotation convention requires only its own small body of pattern-matching rules over the same verb table, and these can be assembled and validated with lightweight scripting at low engineering cost.

\paragraph{Gripper-event detector.}
Per-arm gripper open and close events are extracted from the gripper-command stream by detecting sign-flips between non-zero command values. We prefer this command-based detector to a finger-width threshold because some simulation environments do not uniformly model the kinematic finger response (virtually-attached objects, empty grasps); the command-based detector recovers events that the kinematic detector silently drops.

\paragraph{Verb-to-state-transition mapping.}
SST construction walks the parsed action sequence under the following mapping over the sub-roles defined above. \emph{Grasp-acquire} verbs acquire the named object on the assigned arm; \emph{grasp-release} verbs clear the object from the assigned arm; \emph{grasp-transient} verbs impose a single-subtask hold that is auto-cleared at the subtask boundary; \emph{tool-release} verbs clear the held tool from the assigned arm without acquiring the target. \emph{Persistent tool-use} and \emph{contact} verbs are stateless: they leave the holding state unchanged.

\paragraph{Typed pending-query schema.}
Each open question in $Q_k$ is recorded as a tuple of (query kind, target action index, $\mathtt{requires\_vl}$ flag, free-text reason). Three query kinds arise in practice. An $\mathtt{arm\_assignment}$ query records that the acting arm is ambiguous; it is resolved by L1 when only one arm's gripper transitioned, by L2 when the dataset's coordination metadata disambiguates, and by L3 otherwise. A $\mathtt{same\_object}$ query records that both arms hold same-named objects whose identity is ambiguous (the $\mathsf{D_?}$ configuration of Table~\ref{tab:state-configs}); it is resolved by L2 when instance-level metadata is present, and by L3 otherwise. An $\mathtt{empty\_hand}$ query records that an action is annotated against an empty hand, which is an annotation artefact rather than a perceptual ambiguity; it is resolved deterministically and the $\mathtt{requires\_vl}$ flag is set to false so no VL call is issued.

\paragraph{Online cascade interleaving.}
Construction and cascade are interleaved per SST entry in a snapshot/re-advance pattern. For each new entry $S_k$ we (i)~build the entry without committing state, (ii)~snapshot the current scene-graph context, (iii)~tentatively advance through $A_k$, (iv)~run the cascade on any pending queries in $Q_k$, (v)~restore the snapshot, and (vi)~re-advance through $A_k$ with the resolved values preserved. The resulting after-state $g_k$ becomes the before-state for $S_{k+1}$, so L2 or L3 resolutions at $S_k$ immediately enable L1 inferences at $S_{k+1}$.

\section{VL Prompt Templates}
\label{app:prompts}

This appendix lists the prompts used by the L3 VL layer of the cascade. The cascade issues only two prompt types: \texttt{arm\_assignment} (which arm performs an action) and \texttt{same\_object} (whether two same-named held objects are the same physical instance); the third typed query, \texttt{empty\_hand}, is resolved deterministically without a VL call (Sec.~\ref{sec:resolution}). All templates are applied identically across the 13 VL models; per-model prompt tuning is intentionally not performed, so per-model accuracy in Sec.~\ref{sec:eval-per-model} and Appendix~\ref{app:model-selection} reflects each model's behaviour under a single shared prompt configuration.

\paragraph{System prompt.}
Both query types use the same system prompt:
\begin{quote}\small\ttfamily
You are a robot-manipulation expert analyzing bimanual robot video. Answer strictly in JSON.
\end{quote}

Each query is paired with a 6-camera grid extracted from the subtask's frame range: the start, middle, and end frames of the interval, encoded as a single multi-image input. The grid is constructed identically for every model; provider-specific image-passing conventions are handled by a thin client wrapper that does not modify the prompt text.

\paragraph{\texttt{arm\_assignment} template.}
The prompt is composed at query time from the SST entry's structured fields (action verb, target object, conditions, scene-graph snapshot) and the task-level object inventory. Curly-braced fields are placeholders:
\begin{quote}\small\ttfamily
In these frames from task '\{task\_name\}':\\
Now robot is \{action\_description\}\{conditions\}.\{other\_arm\_context\}\\
Which arm performs this action --- left, right, or coop?\\[2pt]
IMPORTANT: Only these objects exist in the scene: \{object\_list\}. Do NOT identify or reference objects not in this list.\\[2pt]
Respond in JSON format:\\
\{"arm": "left" | "right" | "coop" | "unsure", "confidence": "high" | "medium" | "low", "reasoning": "brief explanation"\}
\end{quote}
The \texttt{other\_arm\_context} field is populated from the entry's scene-graph snapshot \emph{only when the holding state is confirmed} (not \texttt{unsure} and no pending uncertain objects); it states what the non-acting arm is currently holding (e.g.\ ``\emph{Left arm is currently holding bowl.}'') to give the model context for free-arm reasoning. The field is omitted when the scene graph is uncertain, to avoid feeding a guess back into the query.

\paragraph{\texttt{same\_object} template.}
The prompt branches on the candidate arm assignment of the action under test (\texttt{coop}, single-arm, or unspecified):
\begin{quote}\small\ttfamily
\textbf{Branch (a):} both arms acting on the same-named object\\[2pt]
In these frames from task '\{task\_name\}':\\
Both arms are \{action\_description\}.\\
Are they co-grasping the SAME physical \{object\_name\}, or are they each holding a DIFFERENT \{object\_name\} instance?
\end{quote}
\begin{quote}\small\ttfamily
\textbf{Branch (b):} one arm acquiring while the other already holds\\[2pt]
In these frames from task '\{task\_name\}':\\
\{Left|Right\} arm is \{action\_description\}. \{Right|Left\} arm already holds a \{object\_name\}.\\
Is the \{left|right\} arm picking the EXACT SAME physical \{object\_name\} that the \{right|left\} arm holds, or a DIFFERENT one?
\end{quote}
\begin{quote}\small\ttfamily
\textbf{Branch (c):} arm assignment not yet resolved\\[2pt]
In these frames from task '\{task\_name\}':\\
Both arms are reportedly holding a '\{object\_name\}'.\\
Are they holding the SAME physical '\{object\_name\}' cooperatively, or TWO DIFFERENT '\{object\_name\}' instances?
\end{quote}
All three branches share a common tail with the scene-object disclaimer and JSON response schema:
\begin{quote}\small\ttfamily
IMPORTANT: Only these objects exist in the scene: \{object\_list\}. Do NOT identify or reference objects not in this list.\\[2pt]
Respond in JSON format:\\
\{"same\_object": true | false, "confidence": "high" | "medium" | "low", "reasoning": "brief explanation"\}
\end{quote}

\paragraph{Resolution ordering and parsing.}
Within each SST entry, the cascade issues \texttt{same\_object} queries before \texttt{arm\_assignment} queries (Sec.~\ref{sec:resolution}); a confirmed same-object answer can resolve an arm-assignment query deterministically without a further VL call, while the reverse ordering would force the model to pick an arm under same-object ambiguity and frequently produce nonsensical answers. The JSON response is parsed by a tolerant extractor that handles Markdown code blocks, leading or trailing prose, and partial JSON with missing closing braces; runs that fail to produce a valid JSON object after extraction are counted as model-side failures and contribute to the run-to-run instability scores reported in Appendix~\ref{app:model-selection}.

\section{Preliminary Model Selection}
\label{app:model-selection}

The 13-model production pool was assembled in two stages. First, we selected 10 models from a 15-model preliminary panel covering every VL model we could access with stable serving at the time. We evaluated that panel on the same 50 BEHAVIOR-1K tasks and pipeline (3 runs each, both modes, scored against an earlier 99-cell evaluation tier of the annotated reference SSC). The two modes follow the convention used throughout this paper: \emph{VL} denotes vision-language verification only (cascade Level~1 $\to$ Level~3); \emph{AVL} denotes annotations + VL (cascade Level~1 $\to$ Level~2 $\to$ Level~3). Table~\ref{tab:prelim-models} records the per-model accuracy observed under that pipeline and indicates which 10 models were carried forward as the initial production pool. The second stage extended this 10-model pool with three new flagship models (\texttt{gpt-5.5}, \texttt{gemini-3.1-pro-preview}, \texttt{gemini-3.5-flash}) once they became available and passed the same stability checks; these are evaluated under the same pipeline configuration on the same 50 tasks and 3 episodes. The hard-cell definitions of Sec.~\ref{sec:protocol} (138 cells VL trimmed-hard, 56 cells AVL trimmed-hard) are anchored to the original 10-model panel so that per-model and per-trio metrics remain comparable as the pool grows; the three new models are scored on these same cells.

We emphasise that these preliminary numbers reflect what this particular pipeline and prompt configuration produced from each model on a single dataset, and do not necessarily characterise each model's general capability. Several confounds may depress a model's score in this table: (a)~prompt templates, query schema, and JSON output contract were not tuned per model and may suit some families better than others; (b)~BEHAVIOR-1K is a simulated environment whose visual appearance differs from real-world manipulation footage, which can disadvantage models whose pre-training emphasises real-world video; (c)~service availability at the time of the preliminary sweep varied across providers. Selection for the production pool was therefore based on observed \emph{run-to-run stability} and \emph{stable serving at panel scale}, not on the absolute preliminary accuracy.

\begin{table}[h]
\centering
\caption{Preliminary per-model accuracy under our pipeline and prompt configuration on the 99-cell evaluation tier of an earlier annotated reference SSC (15 models, 3 runs, both modes). The right-most column records whether the model was carried forward into the initial 10-model production pool, with a one-phrase reason for non-retention. The three new models added in the second-stage pool extension (\texttt{gpt-5.5}, \texttt{gemini-3.1-pro-preview}, \texttt{gemini-3.5-flash}) are absent from this preliminary panel because they were released after this evaluation cycle. These numbers reflect what this pipeline produced from each model under a single shared prompt set and should not be read as a characterisation of model capability.}
\label{tab:prelim-models}
\small
\begin{tabular}{lrrl}
\toprule
\textbf{Model} & \textbf{VL (\%)} & \textbf{AVL (\%)} & \textbf{Retained?} \\
\midrule
\texttt{gpt-5.4}              & 96.3 & 99.7 & yes \\
\texttt{qwen3-vl-plus}        & 93.9 & 98.3 & yes \\
\texttt{kimi-k2.5}            & 93.9 & 98.7 & yes \\
\texttt{gemini-2.5-pro}       & 93.6 & 96.3 & yes \\
\texttt{doubao-seed-2.0-pro}  & 93.3 & 96.6 & yes \\
\texttt{qwen-vl-max}          & 91.9 & 96.0 & yes \\
\texttt{gpt-4o}               & 90.9 & 91.6 & no (OpenAI family already represented by \texttt{gpt-5.4}) \\
\texttt{qwen3-vl-flash}       & 90.5 & 95.3 & yes \\
\texttt{o4-mini}              & 90.4 & 95.3 & yes \\
\texttt{claude-sonnet-4.6}    & 90.2 & 88.9 & no (run-to-run instability under our prompts) \\
\texttt{gemini-2.5-flash}     & 89.9 & 93.6 & yes \\
\texttt{grok-4}               & 88.9 & 92.9 & no (serving unstable at panel scale) \\
\texttt{gpt-5.4-mini}         & 88.6 & 93.9 & yes \\
\texttt{claude-opus-4.7}      & 83.8 & 87.9 & no (run-to-run instability under our prompts) \\
\texttt{grok-4-fast}          & 75.4 & 80.8 & no (serving unstable at panel scale) \\
\bottomrule
\end{tabular}
\end{table}

The five models excluded from the production pool were dropped on three considerations, all related to pipeline-level stability rather than to inherent model capability.
\textbf{(i)~Run-to-run stability under shared prompts.}
The two Anthropic models (\texttt{claude-sonnet-4.6}, \texttt{claude-opus-4.7}) exhibited the largest VL-only standard deviation across runs (8.5\% for \texttt{claude-sonnet-4.6}), driven primarily by JSON-output compliance failures on a small but consistent set of cases. This is a prompt- and decoding-side artefact that may resolve with model-specific tuning (different JSON schema, repair pass, or system message). However, given the high per-call cost of these models at the panel scale we use here, and the larger scale we envision for downstream applications, model-specific tuning is not an economical option; we therefore excluded these models rather than carry an unstable signal into the production panel.
\textbf{(ii)~Service stability at panel scale.}
The two xAI models (\texttt{grok-4}, \texttt{grok-4-fast}) returned transient service errors during repeated 50-task sweeps. Stable serving is itself a prerequisite at the panel scale we use here and at the larger scale we envision for downstream applications, so we excluded these models on infrastructure grounds rather than on inherent capability.
\textbf{(iii)~Substitutability across a model family.}
\texttt{gpt-4o} was excluded once \texttt{gpt-5.4} became available: we retained one OpenAI flagship plus the mini variant rather than both flagships to control the panel size, without prejudice to \texttt{gpt-4o}'s standalone behaviour.

The retained 10 models span five providers (OpenAI, Google, Alibaba, ByteDance, Moonshot). After the second-stage extension that added \texttt{gpt-5.5} (OpenAI), \texttt{gemini-3.1-pro-preview} (Google), and \texttt{gemini-3.5-flash} (Google), the 13-model production pool covers the same five providers (4 OpenAI + 4 Google + 3 Alibaba + 1 ByteDance + 1 Moonshot), which is sufficient for the provider-diverse ensemble exploration of Sec.~\ref{sec:eval-ensemble}. Anthropic and xAI are absent from the production pool but remain candidates for re-inclusion once model-specific prompt tuning and serving stability issues are revisited.
Finally, we note that the three second-stage additions (\texttt{gpt-5.5}, \texttt{gemini-3.1-pro-preview}, \texttt{gemini-3.5-flash}) are included as soon as they became available following their recent public release, so the final 13-model panel reflects the latest flagship VL models available at the time of writing.

\section{Per-Case Difficulty of the Trimmed-Hard Subset}
\label{app:difficulty}

This appendix breaks down the 138-cell VL trimmed-hard subset (Sec.~\ref{sec:protocol}) by action verb, pool-vote margin, and representative case examples, to characterise where the residual difficulty for VL verifiers lives. The breakdown aggregates across all three episodes.

\paragraph{Verb composition.}
Of the 138 trimmed-hard cells, 134 are panel-disagreement cells and the remaining 4 are retained-ambiguity cells (the pool majority agrees with the human reference but exhibits run-to-run instability). Table~\ref{tab:hard-verbs} reports the verb distribution of the 134 disagreement cases. Disagreements concentrate on \emph{bimanual primitives} rather than on single-arm pick-and-place: \texttt{push} and \texttt{sweep} together account for two thirds of the hard cases, and another quarter come from articulated actions on containers (\texttt{open}, \texttt{close}) and tool-use (\texttt{turn\_on}).

\begin{table}[h]
\centering
\caption{Verb composition of the 134 disagreement cases that drive the trimmed-hard subset. The majority-arm distribution across the 134 cases is \texttt{coop}: 55 (41\%), \texttt{left}: 41 (31\%), \texttt{right}: 38 (28\%); the dominant axis of ambiguity is single-arm versus \texttt{coop}.}
\label{tab:hard-verbs}
\small
\begin{tabular}{lrrl}
\toprule
\textbf{Verb} & \textbf{Disagreements} & \textbf{Share} & \textbf{Typical ambiguity} \\
\midrule
\texttt{push}    & 47 & 35.1\% & one arm bracing while the other pushes, vs.\ \texttt{coop} \\
\texttt{sweep}   & 41 & 30.6\% & broom: left guides, right sweeps, vs.\ \texttt{coop} \\
\texttt{close}   & 16 & 11.9\% & door / cabinet kinematic single-arm vs.\ cooperative \\
\texttt{open}    & 7  & 5.2\%  & dual-handle vs.\ single-handle openings \\
\texttt{place}   & 7  & 5.2\%  & placement onto a large object: \texttt{coop} vs.\ anchor arm \\
\texttt{turn\_on} & 7 & 5.2\%  & one arm holds the appliance, the other operates the control \\
\texttt{pick}    & 6  & 4.5\%  & dual-hold pick of a long or heavy object \\
\textit{other}   & 3  & 2.2\%  & \texttt{hang}, \texttt{tip}, \texttt{release} \\
\midrule
\textbf{Total}   & \textbf{134} & 100\% & \\
\bottomrule
\end{tabular}
\end{table}

\paragraph{Pool-vote margin.}
The pool's vote margin on the 134 disagreements (votes for the majority answer minus the runner-up, out of 10) distributes as: \emph{decisive} (margin $\geq 6$): 93 cases (69\%); \emph{intermediate} (margin 3--5): 29 cases (22\%); \emph{tight} (margin $\leq 2$): 12 cases (9\%). Any reasonable trio resolves the decisive cases correctly by majority vote; the intermediate cases are the main lever for ensemble selection; the tight cases are where the pool itself is nearly split and majority voting may pick either side, and these account for most residual ensemble error.

\paragraph{Representative tight-margin cases.}
Table~\ref{tab:hard-examples} lists five representative tight-margin disagreements drawn from episode~0. Four of the five involve the \texttt{coop} vs single-arm boundary on bimanual tool-use or large-object manipulation; the fifth illustrates a handedness-mirror failure where the visible arm and the reference arm differ because of camera-frame orientation.

\begin{table}[h]
\centering
\caption{Representative tight-margin disagreements from one episode. The \textbf{Cell} column uses the short form \texttt{st<i>.a<j>}, where \texttt{st<i>} is the $i$-th subtask within the episode and \texttt{a<j>} is the $j$-th action within that subtask (zero-indexed); e.g.\ \texttt{st1.a0} is the first action of subtask~1. Pool votes are counts across the 10 single-model verifiers (C = \texttt{coop}, L = \texttt{left}, R = \texttt{right}); \textbf{Ref.}\ is the reference acting arm.}
\label{tab:hard-examples}
\footnotesize
\setlength{\tabcolsep}{4pt}
\begin{tabular}{@{}llp{2.6cm}lcp{4.2cm}@{}}
\toprule
\textbf{Task} & \textbf{Cell} & \textbf{Action} & \textbf{Pool votes} & \textbf{Ref.} & \textbf{Source of difficulty} \\
\midrule
0013 & st1.a0  & open car lid              & C:5 / L:3 / R:2 & C &cooperative open vs.\ split bimanual lift \\
0036 & st46.a0 & sweep floor with broom    & C:6 / R:4       & C &broom needs both hands; active side is right \\
0036 & st48.a0 & sweep floor with broom    & C:5 / R:5       & C &margin 0: fully tied on broom-sweep boundary \\
0029 & st34.a0 & push laptop to bed edge   & R:5 / C:4 / L:1 & R & half the pool sees a bracing left hand \\
0049 & st26.a0 & open fridge right\_door   & R:6 / L:4       & R & handedness mirror: labels read opposite arm \\
\bottomrule
\end{tabular}
\end{table}

\paragraph{Categories of residual difficulty.}
The verb composition and margin distribution together identify three sources of residual difficulty in the trimmed-hard subset.
\textbf{(a)~Bimanual tool-use and large-object manipulation} (push, sweep, place on a large surface): the single-arm vs \texttt{coop} boundary depends on whether the supporting or bracing arm is treated as participating in the action or merely steadying the body. This is a labelling-convention question as much as a perceptual one, and accounts for the bulk of the hard subset.
\textbf{(b)~Articulated contact on containers and controls} (open, close, turn\_on): the acting arm is sometimes visually obvious and sometimes occluded by the container, and coordinated open or close of dual-handle objects can be labelled as either single-arm or \texttt{coop} depending on the dataset's convention.
Categories (a) and (b) account for roughly 80\% of the hard subset and reflect inherent ambiguity in the dataset's labelling convention.

\subsection*{Per-case tiers under the 13-model production pool}
\label{app:difficulty-tiers}

To complement the aggregate breakdown above, we list the hardest cells in the VL trimmed-hard subset under the 13-model production pool of Sec.~\ref{sec:eval-per-model}. Each cell is scored as ``solved'' by a given model when the majority of its three independent runs match the reference label, and ``not solved'' otherwise; the rightmost column reports the count of solving models out of 13. Cases are then tiered by this count: \emph{Tier~1} (zero solvers; universally hard for the production pool) and \emph{Tier~2} (1--2 solvers; near-universally hard) are listed in Table~\ref{tab:tier12}; \emph{Tier~3} (3--6 solvers, i.e.\ at most half the pool; moderately hard) is listed in Table~\ref{tab:tier3}. Cells with seven or more solvers are resolved by any reasonable provider-diverse trio under majority voting and are not detailed individually here.

\begin{table}[h]
\centering
\caption{Tier 1 (top block) and Tier 2 (bottom block) under the 13-model VL production pool: cells solved by 0--2 of the 13 production models. Column codes: DB=\texttt{dbao2p}, G54=\texttt{gpt54}, G5m=\texttt{gpt54mi}, O4=\texttt{o4mi}, Q3=\texttt{qwen3vlp}, Q3f=\texttt{qwen3vlf}, QM=\texttt{qwenvlx}, GP=\texttt{gem25p}, GF=\texttt{gem25f}, KM=\texttt{kimi25}, G55=\texttt{gpt55}, G31=\texttt{gem31p}, G35=\texttt{gem35f} ($^{\star}$ marks the three models added in the second-stage pool extension). \textbf{G}: reference acting arm (C = \texttt{coop}, R = \texttt{right}, L = \texttt{left}, ? = unresolved). \textbf{Src}: \texttt{hum} = human override, \texttt{vlM} = original 10-model VL pool majority. \checkmark = the model solved the cell on a majority of its three runs. The row marked $\dagger$ (\texttt{ep0\,0034 st1 pick poster}) has an unresolved reference arm (G=?) and is recorded for completeness but excluded from the per-model totals in Table~\ref{tab:tier3}.}
\label{tab:tier12}
\scriptsize
\setlength{\tabcolsep}{2pt}
\begin{tabular}{@{}lllll|cccccccccc|ccc|c@{}}
\toprule
\textbf{Ep+Task} & \textbf{ST} & \textbf{Action} & \textbf{G} & \textbf{Src}
& \textbf{DB} & \textbf{G54} & \textbf{G5m} & \textbf{O4} & \textbf{Q3} & \textbf{Q3f} & \textbf{QM} & \textbf{GP} & \textbf{GF} & \textbf{KM}
& \textbf{G55}$^{\star}$ & \textbf{G31}$^{\star}$ & \textbf{G35}$^{\star}$
& \textbf{n/13} \\
\midrule
\multicolumn{19}{l}{\textit{Tier 1: universal failure (0/13).}} \\
ep0 0019 & 2  & open toolbox      & C &hum &            &            &            &            &            &            &            &            &            &            &            &            &            & 0 \\
ep0 0019 & 20 & close toolbox     & C &hum &            &            &            &            &            &            &            &            &            &            &            &            &            & 0 \\
ep0 0029 & 38 & close laptop      & C &hum &            &            &            &            &            &            &            &            &            &            &            &            &            & 0 \\
ep0 0049 & 26 & open fridge       & C &hum &            &            &            &            &            &            &            &            &            &            &            &            &            & 0 \\
ep1 0019 & 1  & push toolbox      & C &hum &            &            &            &            &            &            &            &            &            &            &            &            &            & 0 \\
ep1 0049 & 12 & open fridge       & C &hum &            &            &            &            &            &            &            &            &            &            &            &            &            & 0 \\
\midrule
\multicolumn{19}{l}{\textit{Tier 2: near-universal failure (1/13--2/13).}} \\
\rowcolor{gray!12}
\textcolor{gray!55!black}{ep0 0034$^{\dagger}$} & \textcolor{gray!55!black}{1}  & \textcolor{gray!55!black}{pick poster} & \textcolor{gray!55!black}{?} & \textcolor{gray!55!black}{hum} & \textcolor{gray!55!black}{\checkmark} &            &            &            &            &            &            &            &            &            &            &            &            & \textcolor{gray!55!black}{1} \\
ep1 0029 & 35 & push laptop       & R & hum &            &            &            & \checkmark &            &            &            &            &            &            &            &            &            & 1 \\
ep2 0049 & 22 & open fridge       & C &hum &            &            &            &            &            &            &            & \checkmark &            &            &            &            &            & 1 \\
ep2 0049 & 58 & close oven        & C &hum & \checkmark &            &            &            &            &            &            &            &            &            &            &            &            & 1 \\
ep2 0029 & 37 & close laptop      & C &hum &            &            & \checkmark &            &            &            &            &            &            & \checkmark &            &            &            & 2 \\
\bottomrule
\end{tabular}
\end{table}

\begin{table}[h]
\centering
\caption{Tier 3 under the 13-model VL production pool: cells solved by 3--6 of the 13 production models (at most half the pool). Conventions follow Table~\ref{tab:tier12}. The bottom row reports each model's total solve count across the 20 cells of Tables~\ref{tab:tier12} and~\ref{tab:tier3} combined, excluding the $\dagger$-marked row in Table~\ref{tab:tier12} whose reference arm is unresolved.}
\label{tab:tier3}
\scriptsize
\setlength{\tabcolsep}{2pt}
\begin{tabular}{@{}lllll|cccccccccc|ccc|c@{}}
\toprule
\textbf{Ep+Task} & \textbf{ST} & \textbf{Action} & \textbf{G} & \textbf{Src}
& \textbf{DB} & \textbf{G54} & \textbf{G5m} & \textbf{O4} & \textbf{Q3} & \textbf{Q3f} & \textbf{QM} & \textbf{GP} & \textbf{GF} & \textbf{KM}
& \textbf{G55}$^{\star}$ & \textbf{G31}$^{\star}$ & \textbf{G35}$^{\star}$
& \textbf{n/13} \\
\midrule
ep0 0013 & 1  & open car             & C &vlM & \checkmark & \checkmark & \checkmark &            &            &            & \checkmark & \checkmark &            &            & \checkmark &            &            & 6 \\
ep0 0033 & 11 & push teddy\_bear     & R & hum & \checkmark & \checkmark &            & \checkmark &            &            &            &            &            &            & \checkmark &            & \checkmark & 5 \\
ep0 0049 & 64 & close oven           & C &hum & \checkmark & \checkmark &            &            &            & \checkmark &            & \checkmark &            &            & \checkmark &            & \checkmark & 6 \\
ep1 0049 & 4  & push carving\_knife  & R & vlM & \checkmark & \checkmark &            &            &            &            & \checkmark & \checkmark &            & \checkmark &            &            &            & 5 \\
ep1 0049 & 58 & close oven           & C &hum &            &            & \checkmark &            &            & \checkmark & \checkmark &            &            &            &            & \checkmark & \checkmark & 5 \\
ep2 0016 & 3  & pick storage\_box    & C &hum & \checkmark &            &            & \checkmark &            &            &            &            &            &            & \checkmark & \checkmark & \checkmark & 5 \\
ep2 0016 & 5  & place storage\_box   & C &hum & \checkmark &            &            & \checkmark &            &            &            &            &            &            & \checkmark & \checkmark & \checkmark & 5 \\
ep2 0019 & 2  & open toolbox         & C &hum &            &            & \checkmark &            & \checkmark & \checkmark & \checkmark &            & \checkmark &            &            &            & \checkmark & 6 \\
ep2 0029 & 33 & push laptop          & C &hum &            &            & \checkmark &            &            &            & \checkmark &            &            & \checkmark &            &            &            & 3 \\
ep2 0036 & 1  & push garden\_chair   & C &vlM &            &            & \checkmark &            &            & \checkmark & \checkmark & \checkmark & \checkmark & \checkmark &            &            &            & 6 \\
\midrule
\multicolumn{5}{r|}{\textbf{Total over 20 cells (Tier 1+2+3, excl.\ $\dagger$):}}
& \textbf{7} & \textbf{4} & \textbf{6} & \textbf{4} & \textbf{1} & \textbf{4} & \textbf{6} & \textbf{5} & \textbf{2} & \textbf{4}
& \textbf{5} & \textbf{3} & \textbf{6} & --- \\
\bottomrule
\end{tabular}
\end{table}

\paragraph{Headline.}
On the 20 hardest cells (excluding the $\dagger$-marked row in Table~\ref{tab:tier12}), \texttt{dbao2p} leads at 7/20, followed by \texttt{gpt54mi}, \texttt{qwenvlx}, and the new \texttt{gem35f} at 6/20. The new \texttt{gpt55} solves 5/20 and \texttt{gem31p} solves 3/20, neither approaching \texttt{dbao2p}'s coverage on the residual VL-hard cells. \texttt{qwen3vlp} scores only 1/20 under VL --- the worst pool member on these cases --- despite topping the AVL trimmed-hard column in Appendix~\ref{app:avl-ablation}: its strength is concentrated on easier cases after the annotation layer resolves, not on the residual VL-hard cells. All six Tier-1 cells involve coordinated bimanual open, close, or push actions on large containers such as toolboxes, fridges, ovens, and laptops. The failure mode persists after adding the three new flagship models: they add Tier-3 coverage (the \texttt{storage\_box} pair, \texttt{push teddy\_bear}, and one \texttt{open car} cell) but solve no Tier-1 cell. The residual cohort is the source of the $\sim$4--5\% gap that the strongest trios cannot close on the 138-cell subset, and is the most concrete improvement target for future verifier work.

\section{AVL Mode Ablation: What Annotation Lookup Adds}
\label{app:avl-ablation}

When the dataset supplies structured metadata that can disambiguate same-object identity and bimanual coordination (Sec.~\ref{sec:resolution}), the cascade's optional Level~2 stage consumes it. This appendix quantifies what L2 buys on BEHAVIOR-1K relative to the VL-only mode of Sec.~\ref{sec:eval-per-model}, treating AVL as an ablation of the main VL-only reporting. We define the \emph{AVL trimmed-hard} subset of 56 cells analogously to the 138-cell VL trimmed-hard set (Sec.~\ref{sec:protocol}): cells on which the 13-model panel under AVL mode does not produce the correct label by majority vote, retaining only those that remain hard once Level-2 annotation lookup is enabled. Table~\ref{tab:per-model-avl} repeats the VL-only rows from Table~\ref{tab:per-model}, then adds two AVL views: performance on the 56-cell AVL-hard subset, and \emph{cross-mode} performance on the same 138 VL-hard cells (the only apples-to-apples comparison between the two modes).

\begin{table}[h]
\centering
\caption{Per-model accuracy under both operating modes, plus cross-mode AVL scored on the 138-cell VL-hard subset. Models are listed in the same left-to-right order as Table~\ref{tab:per-model} (VL trimmed-hard ranking); the first two rows are copied from Table~\ref{tab:per-model} for ease of comparison. The bottom row reports $\Delta$ (AVL on VL-hard $-$ VL Hard/138) in percentage points, the apples-to-apples gain from enabling Level~2 on each model's same hard cells. Best per row in bold.}
\label{tab:per-model-avl}
\scriptsize
\setlength{\tabcolsep}{0.5pt}
\begin{tabular}{@{}l|ccccccccccccc@{}}
\toprule
\textbf{Model} & \texttt{gpt55} & \texttt{gem35f} & \texttt{gpt54} & \texttt{gem31p} & \texttt{dbao2p} & \texttt{gem25p} & \texttt{qwen3vlf} & \texttt{kimi25} & \texttt{gem25f} & \texttt{qwenvlx} & \texttt{gpt54mi} & \texttt{o4mi} & \texttt{qwen3vlp} \\
\midrule
VL Overall (\%)         & \textbf{99.19} & 99.15 & 99.11 & 99.07 & 99.02 & 99.02 & 98.90 & 98.73 & 98.51 & 98.13 & 98.01 & 98.01 & 97.84 \\
VL Hard/138 (\%)        & \textbf{86.23} & 85.51 & 84.78 & 84.78 & 83.33 & 83.33 & 81.16 & 78.26 & 74.64 & 68.12 & 65.94 & 65.94 & 65.94 \\
\midrule
AVL Overall (\%)        & \textbf{99.70} & \textbf{99.70} & 99.62 & 99.62 & 99.32 & 99.32 & 99.32 & 99.36 & 99.19 & 99.32 & 99.07 & 98.98 & \textbf{99.70} \\
AVL Hard/56 (\%)        & \textbf{87.50} & \textbf{87.50} & 83.93 & 85.71 & 71.43 & 71.43 & 71.43 & 73.21 & 66.07 & 71.43 & 60.71 & 57.14 & \textbf{87.50} \\
\midrule
AVL on VL-hard/138 (\%) & \textbf{95.65} & \textbf{95.65} & 94.93 & \textbf{95.65} & 91.30 & 90.58 & 89.86 & 90.58 & 87.68 & 89.86 & 85.51 & 85.51 & \textbf{95.65} \\
$\Delta$ vs VL (pp)     & $+9.4$ & $+10.1$ & $+10.2$ & $+10.9$ & $+8.0$ & $+7.3$ & $+8.7$ & $+12.3$ & $+13.0$ & $+21.7$ & $+19.6$ & $+19.6$ & $\mathbf{+29.7}$ \\
\bottomrule
\end{tabular}
\end{table}

\paragraph{What annotation lookup buys.}
The cross-mode row of Table~\ref{tab:per-model-avl} makes the comparison fair: AVL predictions are scored on the same 138 cells used by the VL row above it, so the row immediately above is the apples-to-apples gain from enabling Level~2. The strongest single model \texttt{gpt55} climbs from 86.23\,\% under VL to 95.65\,\% under AVL --- a lift of $+9.4$\,pp. The model ordering reshuffles between modes: \texttt{qwen3vlp} sits at the bottom of the VL column (65.94\,\%) but jumps to a three-way tie at the top of the AVL trimmed-hard column (87.50\,\%) alongside \texttt{gpt55} and \texttt{gem35f}, reflecting a model-specific bias in same-object reasoning that the L2 lookup absorbs when the dataset's metadata is reliable. The annotation-driven swing on the shared 138-cell subset (last row of Table~\ref{tab:per-model-avl}) is largest for \texttt{qwen3vlp} ($+29.7$\,pp) and \texttt{qwenvlx} ($+21.7$\,pp), and smallest for the newer flagships (\texttt{gpt55}: $+9.4$\,pp; \texttt{gem35f}: $+10.1$\,pp), suggesting that the newer models already extract much of the signal from raw VL grids that the annotation layer would otherwise contribute.

\paragraph{q3vlp asymmetry: verification, not reasoning.}
\texttt{qwen3vlp}'s extreme VL$\to$AVL gap ($65.94\% \to 95.65\%$ on the 138-cell subset, $+29.7$\,pp) reflects a difference in what the two modes actually test, rather than a step-change in model capability.

\emph{Mode mechanism.} The model's per-action stochasticity is $\sim\!1\%$ under VL but below $0.1\%$ under AVL: L2 annotation overrides dominate the AVL decision path, so the model is effectively asked ``do you accept this annotation?'' rather than ``which arm performs this action?''. The VL flip pattern is approximately symmetric (left$\to$right balances right$\to$left across episodes), consistent with the model guessing on cells it cannot confidently resolve. The 56-cell AVL trimmed-hard subset is also the residual after annotation has already settled the easier ambiguities, so a high AVL rank on those 56 cells is a strictly easier benchmark than a high VL rank on the 138-cell hard subset.

\emph{Practical implication.} \texttt{qwen3vlp}'s AVL strength reflects verification behaviour rather than fresh VL reasoning on hard frames; it should be paired with a stronger VL reasoner when included in an ensemble.

\paragraph{Worked-example trio under AVL.}
The worked-example trio \texttt{gpt54 + qwenvlx + gem35f} (Sec.~\ref{sec:eval-ensemble}) reaches $89.29\%$ ($50/56$) on AVL trimmed-hard with weakest-member accuracy $71.43\%$, making it the only top-15 VL trio that also appears in the top-15 of the AVL leaderboard (Appendix~\ref{app:ensemble}); this is why it was chosen as the worked example in the main text.

\paragraph{Caveat on the lift.}
The AVL gain is opportunistic and depends on whether the dataset ships with structured metadata that maps reliably onto same-object identity and bimanual coordination. On BEHAVIOR-1K each of \texttt{skill\_type}, \texttt{memory\_prefix}, and \texttt{spatial\_prefix} is unevenly populated (Sec.~\ref{sec:protocol}), so the AVL numbers in this appendix are an upper bound on what L2 can recover from the dataset metadata that is actually present. Datasets without such metadata fall through to L1\,$\to$\,L3 and the VL-only column of Table~\ref{tab:per-model} applies.

\section{Ensemble Exploration: Top Provider-Diverse Trios}
\label{app:ensemble}

This appendix lists the top-15 three-model ensembles from the 13-model production pool, ranked by accuracy on the 138-cell VL trimmed-hard subset. We use the trio leaderboard to characterise the ensemble landscape rather than to select a single optimal trio; the gaps between adjacent ranks are typically a single cell out of 138, which sits within the run-to-run variation we observe per (model, cell). For each trio we annotate whether it is provider-diverse (at most one model per provider, under the convention used by Sec.~\ref{sec:eval-ensemble}) and whether it also appears in the top-15 of the AVL trimmed-hard leaderboard (56-cell denominator).

\begin{table}[h]
\centering
\caption{Top-15 three-model ensembles on the 138-cell VL trimmed-hard subset. \textbf{Div}: \texttt{Y} = 3-provider-diverse (at most one model per provider), \texttt{N} = two of the three members share a provider. \textbf{AVL\,T15}: \texttt{Y} = trio appears in the top-15 of the AVL trimmed-hard leaderboard (56-cell denominator), \texttt{N} = not present. Provider abbreviations: B=ByteDance, O=OpenAI, A=Alibaba, G=Google, M=Moonshot.}
\label{tab:trio-leaderboard}
\small
\setlength{\tabcolsep}{4pt}
\begin{tabular}{rlllcc}
\toprule
\textbf{Rank} & \textbf{Trio} & \textbf{Providers} & \textbf{VL Hard (138)} & \textbf{Div} & \textbf{AVL\,T15} \\
\midrule
1  & \texttt{gpt54mi + gpt55 + gem35f}            & O+O+G & 89.86\% (124/138) & N & N \\
2  & \texttt{dbao2p + qwen3vlf + gem35f}          & B+A+G & 89.13\% (123/138) & Y & N \\
2  & \texttt{dbao2p + qwenvlx + gem35f}           & B+A+G & 89.13\% (123/138) & Y & N \\
4  & \texttt{dbao2p + gpt54mi + gem35f}           & B+O+G & 88.41\% (122/138) & Y & N \\
4  & \texttt{gpt54 + qwenvlx + gem35f} $^\star$   & O+A+G & 88.41\% (122/138) & Y & Y \\
4  & \texttt{qwen3vlf + gpt55 + gem35f}           & A+O+G & 88.41\% (122/138) & Y & N \\
4  & \texttt{qwenvlx + gpt55 + gem35f}            & A+O+G & 88.41\% (122/138) & Y & N \\
8  & \texttt{dbao2p + gpt54 + gem31p}             & B+O+G & 87.68\% (121/138) & Y & N \\
8  & \texttt{dbao2p + qwenvlx + gpt55}            & B+A+O & 87.68\% (121/138) & Y & N \\
8  & \texttt{dbao2p + qwenvlx + gem31p}           & B+A+G & 87.68\% (121/138) & Y & N \\
8  & \texttt{gpt54 + qwen3vlf + gem25p}           & O+A+G & 87.68\% (121/138) & Y & N \\
8  & \texttt{gpt54 + gem25p + gpt55}              & O+G+O & 87.68\% (121/138) & N & N \\
8  & \texttt{gpt54 + gpt55 + gem31p}              & O+O+G & 87.68\% (121/138) & N & N \\
8  & \texttt{gpt54 + gpt55 + gem35f}              & O+O+G & 87.68\% (121/138) & N & N \\
8  & \texttt{gpt54 + gem31p + gem35f}             & O+G+G & 87.68\% (121/138) & N & N \\
\bottomrule
\end{tabular}
\end{table}

\noindent $^\star$ Worked example used in Sec.~\ref{sec:eval-ensemble}, chosen because it is the only top-15 VL trio that also appears in the top-15 of the AVL leaderboard, providing both-mode coverage from a single trio.

\paragraph{Three observations.}
First, the gap between the leading trio (89.86\%) and rank 15 (87.68\%) is 3 cells out of 138, comparable to the variation a single model exhibits across reruns on the same cells, so the leaderboard reads more naturally as a top-cluster than as a strict ranking.
Second, 10 of the 15 top-VL trios are 3-provider-diverse and 5 are not; the non-diverse trios concentrate two OpenAI or two Google models, reflecting the strong individual performance of the four new flagship members (\texttt{gpt55}, \texttt{gem31p}, \texttt{gem35f}, and the established \texttt{gpt54}).
Third, the cross-mode overlap is sparse: of the 15 trios listed here, only \texttt{gpt54 + qwenvlx + gem35f} also appears in the top-15 of the AVL trimmed-hard leaderboard. Trio selection therefore does not transfer cleanly between modes: a trio optimised for VL-only accuracy is unlikely to be optimal under AVL, and vice versa. This is an additional reason why we treat the worked example as a representative reference rather than as a canonical recommendation; a definitive ensemble selection would require multi-dataset, larger-run-count validation that lies outside the scope of this work.

\end{document}